\documentclass[11pt, copyright, logo, secondlogo]{template/lark}

\usepackage[authoryear, sort&compress, round]{natbib}
\usepackage{bbm}
\usepackage{enumitem}
\setlist{nosep}
\usepackage{xspace}
\usepackage{graphicx}
\usepackage{tabularx}
\usepackage{booktabs}
\usepackage{subcaption}
\usepackage{xcolor}
\usepackage{colortbl}
\usepackage{hyperref}
\usepackage{verbatim}
\usepackage{float}
\usepackage{cleveref}
\usepackage{wrapfig}
\usepackage{makecell}
\usepackage{multirow}
\usepackage{fancyvrb}
\usepackage[normalem]{ulem}
\usepackage{url}
\usepackage[most]{tcolorbox}
\usepackage{listings}
\usepackage{algorithm}
\usepackage{algpseudocode}
\usepackage{fontawesome5}

\usepackage{amsmath,amsfonts,bm}

\def\eqref#1{equation~\ref{#1}}

\def\1{\bm{1}}

\DeclareMathAlphabet{\mathsfit}{\encodingdefault}{\sfdefault}{m}{sl}
\SetMathAlphabet{\mathsfit}{bold}{\encodingdefault}{\sfdefault}{bx}{n}

\newcommand{\Cov}{\mathrm{Cov}}

\definecolor{darkgreen}{RGB}{0,128,0}
\newcommand{\our}{\textsc{EnvFactory}\xspace}

\title{\our: Scaling Tool-Use Agents via Executable Environments Synthesis and Robust RL}

\author[1,*]{Minrui Xu}
\author[1,*]{Zilin Wang}
\author[1]{Mengyi Deng}
\author[1]{Zhiwei Li}
\author[1]{Zhicheng Yang}
\author[1]{Xiao Zhu}
\author[3]{Yinhong Liu}
\author[4]{Boyu Zhu}
\author[1]{Baiyu Huang}
\author[1]{Chao Chen}
\author[2]{Heyuan Deng}
\author[2]{Fei Mi}
\author[2]{Lifeng Shang}
\author[2,$\dagger$]{Xingshan Zeng}
\author[1,5$\dagger$]{Zhijiang Guo}

\affil[1]{LARK, HKUST (GZ)}
\affil[2]{Huawei Technologies Co., Ltd}
\affil[3]{University of Cambridge}
\affil[4]{UCL}
\affil[5]{HKUST}

\correspondingauthor{Zhijiang Guo(zhijiangguo@hkust-gz.edu.cn), Xingshan Zeng(zeng.xingshan@huawei.com) }

\githubpage{https://github.com/LARK-AI-Lab/EnvFactory}

\begin{abstract}
Equipping LLMs with tool-use capabilities via Agentic Reinforcement Learning (Agentic RL) is bottlenecked by two challenges: the lack of scalable, robust execution environments and the scarcity of realistic training data that captures implicit human reasoning. Existing approaches depend on costly real-world APIs, hallucination-prone LLM simulators, or synthetic environments that are often single-turn or depend on pre-collected documents. Moreover, synthetic trajectories are frequently over-specified, resembling instruction sequences rather than natural human intents, reducing their effectiveness for RL training. We introduce \textbf{\our}, a fully automated framework that addresses both challenges. \our autonomously explores and verifies stateful, executable tool environments from authentic resources, and synthesizes natural multi-turn trajectories through topology-aware sampling and calibrated refinement, producing grounded queries with implicit intents. Using only 85 verified environments across 7 domains, \our generates 2,575 SFT and RL trajectories. Despite using significantly fewer environments than prior work, which are often 5 times more, \our achieves superior training efficiency and downstream performance, improving Qwen3-series models by up to \textbf{+15\%} on BFCLv3, \textbf{+8.6\%} on MCP-Atlas, and \textbf{+6\%} on conversational benchmarks including $\tau^2$-Bench and VitaBench. By fully automating both environment construction and trajectory synthesis, \our provides a scalable, extensible, and robust foundation for Agentic RL.
\end{abstract}

\begin{document}
\maketitle

\section{Introduction}

Equipping Large Language Models (LLMs) with tool-use capabilities has significantly expanded the frontier of AI agents~\citep{toollearningsurvey,llmagentsurvey2025}. Interacting with external tools enables real-time information retrieval, precise computation, and complex system orchestration. Early approaches~\citep{toolmind2025,toucan2025} typically rely on \textit{supervised fine-tuning} (SFT) to teach tool-calling formats and interaction patterns, while more work explores \textit{agentic reinforcement learning} (Agentic RL), where agents acquire tool-use policies through trial-and-error interactions with users and executable environments~\citep{FCviaRL2025,searchr12025,retool2025}. Such frameworks typically involve three key components: agents, environments, and users. The interplay between these components is critical for learning effective tool-use abilities. 

The effectiveness of Agentic RL ultimately hinges on two core factors: \textbf{environments} and \textbf{data}. Scalable and executable environments must faithfully capture real-world interaction dynamics while ensuring low-latency and stable execution. Meanwhile, realistic and verified tool-use data, which reflects contextual ambiguity and implicit reasoning, are essential for improving generalization and providing reliable reward signals for stable policy optimization.

However, existing approaches fall short on either fronts. From the environment perspective, prior methods generally fall into three categories. (1) \emph{Production environments}~\citep{toolllm2023,stabletoolbench2025,toucan2025,hardgen2026}, such as real-world APIs or MCPs, provide authentic execution, but remain costly to scale and destabilize RL training due to potential network latency. (2) \emph{Simulated environments}~\citep{simulatingenvironments2025,word2word2026,scalingagentlearningexperience2025} use LLMs to emulate tool behavior, enabling rapid prototyping but often suffering from hallucination, which makes RL training difficult to generalize in real-world application~\citep{languagemodelshallucinate2025,languagemodelsservetextbased2024}. (3) \emph{Synthetic environments} reconstruct tools through sandboxed code, offering a balance between realism and scalability~\citep{autoforge2025,agentscaler2025}. However, existing synthetic methods exhibit several key limitations: some approaches rely solely on stateless environments~\citep{proceduralenvironment2025,feedbackdriven2026}, while others depend on pre-collected documents, which limits their generalization to unseen tool ecosystems~\citep{autoforge2025,agentscaler2025}.

Another gap exists on the data side. In real-world, user requests are often concise and implicit, requiring agents to perform logical inference and contextual reasoning. Capturing such interaction patterns is crucial, as they faithfully reflect real-world usage while introducing richer decision-making challenges for agent training. However existing synthetic trajectories are commonly over-specified to ensure pass rate, explicitly enumerating task requirements and reasoning steps~\citep{magnet2025,toucan2025}. Consequently, these trajectories resemble rigid ``instruction lists'' rather than natural human intents, limiting both their realism and value for training agentic decision-making.

To address these limitations, we propose \textbf{\our}, a fully automated framework that unifies robust environment construction and realistic trajectory generation with topology-aware graph-based guidance. At the environment level, \our autonomously proposes diverse tool-use scenarios and explores authentic online resources, enabling scalable expansion to previously unseen tool ecosystems while preserving strong fidelity to real-world usage. Based on these structured proposals, \our automatically constructs stateful databases and executable tool interfaces, followed by rigorous verification and iterative refinement to ensure robustness. This fully automated pipeline enables the scalable creation of diverse, low-latency, and reliable environments for Agentic RL. 

At the data level, \our addresses the realism gap in existing synthetic trajectories by two strategies: First, a topology-aware sampling strategy recursively resolves logical dependencies during sampling, ensuring that the guided tools form a coherent logical foundation for query generation. Second, a calibrated refining stage injects realistic human communication patterns—including implicit intents and ambiguity—into the generated queries, transforming the rigid ``instruction lists'' into natural human requests.

\begin{figure*}[t]
    \centering
    \includegraphics[width=0.95\textwidth]{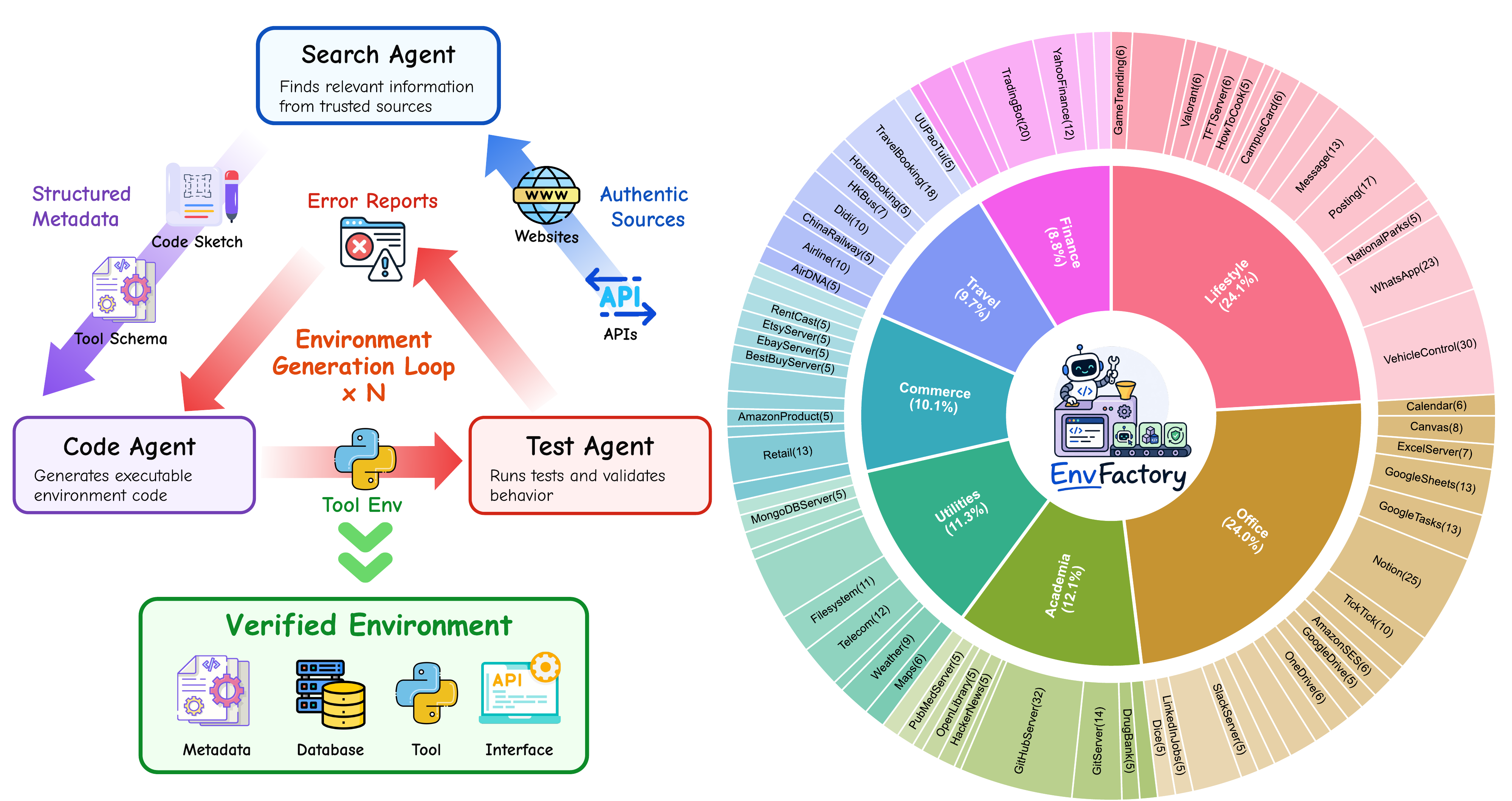}
    \caption{\small The left figure presents an overview of \textit{EnvGen}: the Search Agent autonomously proposes and searches for authentic sources; the Code Agent implements the database and code using feedback from the Test Agent; and the Test Agent generates test cases and error reports. The collaboration between three agents construct diverse, verified environments. The right figure displays a sunburst plot of environments , with the inner ring indicating the proportion of each domain they belongs to and the outer ring showing the number of tools for each environment.}
    \label{fig:env_gen}
    \vspace{-3mm}
\end{figure*}

Using \our, we construct 85 verified environments comprising 842 tools across diverse domains, including \textit{commerce, finance, travel, office, lifestyle, research, and utilities}, as illustrated in Figure~\ref{fig:env_gen}. Building on these environments, we synthesize 1,622 SFT and 953 RL multi-turn, multi-step trajectories for post-training. Despite using significantly fewer environments than concurrent work~\citep{envscaler2026,awm2026}, which are often 5 times more, \our achieves higher training efficiency and stronger downstream performance, improving Qwen3-series models by up to \textbf{15\%} on BFCLv3, \textbf{8.6\%} on the real-world MCP benchmark MCP-Atlas, and \textbf{6\%} on conversational benchmarks, including $\tau^2$-Bench and VitaBench. We summarize our contributions as follow:
\begin{itemize}[leftmargin=1em, itemsep=1pt, topsep=1pt, parsep=0pt, partopsep=0pt]
    \item We propose \textbf{\our}, a unified autonomous pipeline for scaling diverse, executable tool environments and synthesizing realistic, verified trajectories for both SFT and RL training.
    \item We introduce a novel topology-aware sampling algorithm that recursively resolves tool dependencies and synthesizes coherent, natural multi-turn trajectories with implicit intents.
    \item Extensive experiments highlight the data efficiency of \our and its effectiveness for training agents in complex tool-use environments.
\end{itemize}

\section{Related Work}
\label{sec:related_work}
\noindent\textbf{Environment Scaling for Tool Agents.}
The tool-augmented LLM agents is deeply tied to the quality of environments. Existing environment construction strategies fall into three paradigms. \textbf{Production environments} employ real-world APIs~\citep{toolllm2023} and MCP servers~\citep{toucan2025} to provide authentic execution. However, they are expensive to scale and suffer from network latency, which destabilizes RL training. \textbf{Simulated environments} leverage LLMs to emulate tool behavior and state dynamics, enabling rapid prototyping~\citep{simulatingenvironments2025,word2word2026,scalingagentlearningexperience2025}. However, they are prone to hallucination and introduce both expense and instability, making them difficult to generalize to real-world application~\citep{languagemodelshallucinate2025,languagemodelsservetextbased2024}. \textbf{Synthetic environments} reconstruct tools and databases through sandbox code generation, offering a practical compromise between realism, scalability, and training stability~\citep{agentscaler2025,autoforge2025,awm2026,envscaler2026,hardgen2026}. However, AutoForge~\citep{autoforge2025} and AgentScaler~\citep{agentscaler2025} rely on pre-collected tools or documentation, EnvScaler~\citep{envscaler2026} builds on existing task sets, and AWM~\citep{awm2026} starts from abstract scenario seeds, rather than directly recovering real online tool ecosystems. In contrast, \our autonomously discovers tools from authentic online resources, eliminating reliance on pre-curated specifications. By automatically constructing stateful databases and executable tool interfaces with rigorous verification, \our delivers scalable, robust environments grounded in real-world tool ecosystems.

\noindent\textbf{Dependency Tool Graph.}
Sequential tool-use queries often involve strong dependencies among tools, making it challenging for LLMs to generate realistic trajectories directly~\citep{trajectorybench2025,sitgraph2026,gap2025}. A common solution constructs a directed dependency graph over available tools and samples valid sequences via graph traversal. Tool graphs are typically built using either (1) semantic similarity matching between tool parameters and descriptions~\citep{gtool2025,toolflow2025}, which is efficient but may miss implicit logical relationships; or (2) LLM-based reasoning to infer dependencies~\citep{agentscaler2025}, which is more flexible but computationally expensive and potentially inconsistent. Once constructed, these graphs are commonly traversed via naive random walks~\citep{magnet2025,sog2025}, which often fail to fully resolve dependencies---particularly when a tool requires outputs from multiple preceding tools. In contrast, our approach combines semantic matching with LLM-augmented refinement for graph construction, and introduces a topology-aware sampling strategy that recursively resolves unsatisfied input dependencies before tool selection. More related work is discussed at Appendix~\ref{appendix:related}.

\section{Method}
\label{Method}
\subsection{Problem Setup: Tool Agentic Interaction}
\label{sec:problem_setup}
We define the tool agentic interaction between users, agents, and environments as follow:

\noindent\textbf{Environments} ($\mathcal{E}$). Let $\mathcal{E}$ denote the set of available tool environments. Each environment $e \in \mathcal{E}$ is defined as $e=(m,\mathcal{D},\pi,\mathcal{V}_e)$, where $m$ denotes environment metadata (e.g., descriptions, tool definitions, and tool schemas), $\mathcal{D}$ is the stateful database schema specifying the underlying environment state, $\pi$ is the executable Python implementation, and $\mathcal{V}_e$ is the tool interface exposed to the agent (e.g., tool names, descriptions, and parameter specifications), use MCP~\citep{mcp2024} by default.

\noindent\textbf{Tools} ($\mathcal{V}$). Each environment $e \in \mathcal{E}$ exposes a tool interface $\mathcal{V}_e$, and the global toolset is defined as $\mathcal{V}=\bigcup_{e\in\mathcal{E}}\mathcal{V}_e$. Each tool $v\in\mathcal{V}$ is associated with an input space $\mathcal{I}(v)$ and an output space $\mathcal{O}(v)$.

\noindent\textbf{Agent}. At each step, the agent observes the user message or tool execution results, and chooses either to invoke tools from $\mathcal{V}$ or to emit a natural-language response to the user.

\noindent\textbf{User}. When receiving the agent's message, the user may provide additional information, clarify the agent's questions, or perform instructed actions.

For each turn, the interaction continues until either a predefined maximum number of steps is reached or the user proactively terminates the conversation by emitting a stop token.

\noindent\textbf{Overview}. To synthesize high-quality tool agentic interaction trajectories, \our first constructs environments autonomously using \textit{EnvGen}, yielding an executable environment set $\mathcal{E}$ and corresponding tool set $\mathcal{V}$. Using $\mathcal{V}$, we build a dependency tool graph $G$ that captures relationships among tools. Leveraging $G$, we then employ a topology-aware sampling strategy to randomly sample an ordered list of tools $\tau=[v_1,...,v_n]$, which serves as the backbone for synthesizing multi-step, multi-turn tool agentic interaction trajectories using \textit{QueryGen}.
\subsection{Environment Construction}
\noindent\textbf{Overview.}
Given an empty set of environment $\mathcal{E}=\emptyset$, \textit{EnvGen} fully automates the construction of a new environment $e_{\text{new}} = (m, \mathcal{D}, \pi, \mathcal{V}_{e_{\text{new}}}) \notin \mathcal{E}$ by generating diverse proposals, retrieving authentic sources, and iteratively implementing, executing, and revising to ensure a stable training environment, as shown in Figure~\ref{fig:env_gen}. The environment pool is subsequently augmented as $\mathcal{E} \leftarrow \mathcal{E} \cup \{e_{\text{new}}\}$.

\noindent\textbf{Proposal and Sketch.}
Instead of drafting environments from static documents, our Search Agent plans and sketches candidate environments with authentic external sources. The agent analyzes the current environments $\mathcal{E}$ to identify coverage gaps and retrieves source-grounded, broadly applicable functionalities---such as API documentation, technical reports, and usage examples---to inform environment designs. For each selected candidate, it then produces structured metadata $m$, including environment descriptions, tool definitions, and tool schemas, which serve as a blueprint for constructing $e_{\text{new}}$. By grounding environment proposals in authentic and widely applicable functionalities, this stage promotes the diversity, authenticity, and scalability of the generated environments.

\noindent\textbf{Database Modeling.}
\label{sec:database_modeling}
Given metadata $m$, a Code Agent derives a stateful database schema $\mathcal{D}$ that captures the entities, relationships, and mutable states needed to support the environment's functionalities. Tool parameters, intermediate states, and persistent records 
are formalized as Pydantic schemas with standardized serialization interfaces for loading and dumping states. This design ensures clean session isolation and reproducible execution across training rollouts.

\noindent\textbf{Code Implementation.}
Conditioned on $m$ and $\mathcal{D}$, the Code Agent implements executable Python code $\pi$ for each tool, ensuring consistency with the specified functionality, constraints, and schema definitions. The implementations are then wrapped into a standardized tool interface $\mathcal{V}_{e_{\text{new}}}$ (e.g., MCP), exposing well-defined tool names, descriptions, and parameter specifications to agents.

\noindent\textbf{Revision Loop.}
After constructing $\mathcal{D}$, $\pi$, and $\mathcal{V}_{e}$, a Test Agent creates unit test cases and validates the environment against four criteria: (1) tool interfaces are consistent with metadata $m$; (2) tools import and execute successfully; (3) execution results match expected behavior; and (4) database states transition correctly after tool invocation. Upon failure, the Test Agent produces a structured error report that localizes the source (e.g., implementation logic) and provides revision suggestions. The Code Agent then updates the corresponding component and rebuilds the environment. This iterative validation-and-revision loop continues until all tests pass or a maximum revision budget is reached. The final verified environment $e_{\text{new}}=(m,\mathcal{D},\pi,\mathcal{V}_{e_{\text{new}}})$ is cross-validated across all components, ensuring stable and reproducible execution during RL training.
\subsection{Dependency Tool Graph}
\subsubsection{Tool Graph Construction}
\label{sec:tool_graph_construction}

\begin{figure*}[t]
    \centering
    \includegraphics[width=0.99\textwidth]{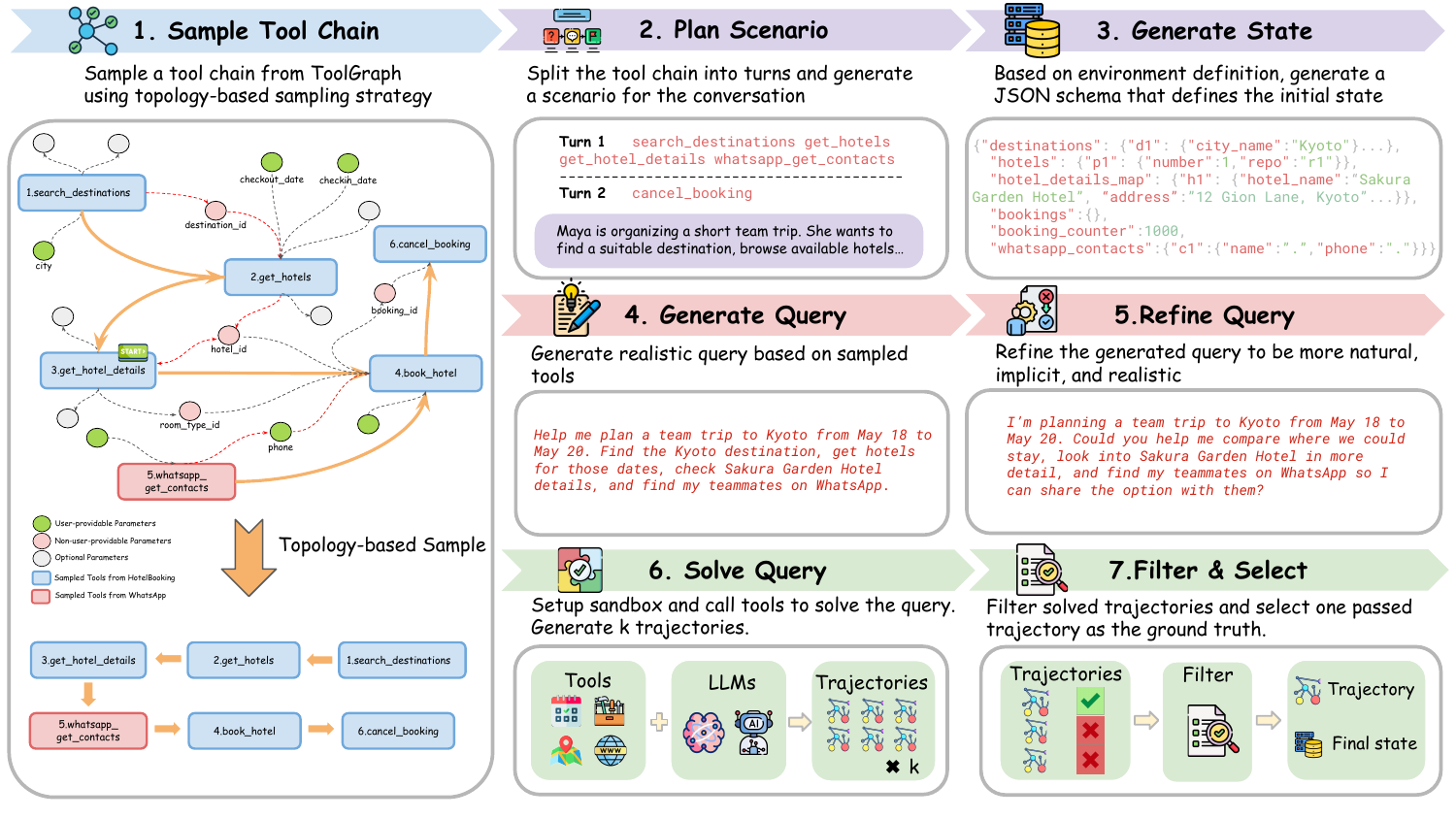}
    \vspace{-2mm}
    \caption{\small  The overall framework of \textit{QueryGen}: Part 1 illustrates the topology-aware sampling strategy, highlighting its non-linear nature, while Parts 2–7 detail the step-by-step synthesis of queries.}
    \label{fig:query_gen}
    \vspace{-2mm}
\end{figure*}

We construct a tool dependency graph $G = (\mathcal{V}, E)$ using semantic matching to capture the nonlinear relationships between tools. However, relying solely on semantic similarity is insufficient to model all logical dependencies. For instance, tools without input or output parameters and tools that belong to the same functional group despite differing signatures may not be adequately represented. To address these limitations, we propose a fine-grained method that models both tools and their parameters as nodes in $G$, resulting in a graph that is more semantically coherent and logically sound.

\noindent\textbf{Step 1: Semantic Parameter Matching}.
Using the \texttt{BAAI/bge-m3} embedding model~\citep{m3embedding2025}, we encode all input and output parameters of every tool. For any pair of tools $(v_i, v_j) \in \mathcal{V} \times \mathcal{V}$, we compute the cosine similarity between the embeddings of every output parameter $p_o \in \mathcal{O}(v_i)$ and every input parameter $p_i \in \mathcal{I}(v_j)$. If any such similarity exceeds a preset threshold, we add a directed edge $(v_i \rightarrow v_j)$ to $G$, indicating that $v_j$ may consume outputs produced by $v_i$.

\noindent\textbf{Step 2: Logical Dependency Refinement}.
For each environment $e\in\mathcal{E}$, we further prompt a LLM to analyze the tools in $\mathcal{V}_e$, identify missing logical dependencies and prune spurious edges introduced by semantic matching. This step is essential because parameter-less tools will be otherwise isolated. For example, in the \texttt{Notion} environment, the tool \text{delete\_all\_notes} accepts no input parameters and returns no output parameters; without further refinement, it would be disconnected from the graph.

\subsubsection{Topology-Aware Sampling}
\label{sec:topology_sampling}
Leveraging the tool graph $G$, we sample a tool sequence $\tau = [v_1, v_2, \ldots, v_n]$ to guide the synthesis of realistic tool-use queries. However, two challenges bottleneck this process. 
First, vanilla sampling strategies such as random walk only capture sequential logic, whereas real-world scenarios often demand non-linear reasoning patterns. 
Second, synthesizing natural user queries from sampled tool chains requires that missing input parameters be realistically satisfiable---either provided explicitly by the user or derived from the outputs of preceding tools in the chain. 
To address both challenges, we enforce the following sampling constraint: \textit{All required input parameters of a sampled tool must be either externally provided by the human user or internally derived from the outputs of previously sampled tools.} Figure~\ref{fig:query_gen} shows an example of topology-aware sampling strategy.

\noindent\textbf{Identify Internal and External Parameters.}
\label{sec:internal_external_param}
We employ an LLM to classify each input parameter as either \textit{external} or \textit{internal}. 
External parameters (e.g., \texttt{city}, \texttt{name}) require explicit provision from an external source such as a human user. 
In contrast, internal parameters (e.g., \texttt{hotel\_id} for \texttt{book\_hotel}) depend on the outputs of preceding tool calls (e.g., \texttt{get\_hotel\_list}), representing internal system states that users are unlikely to know or recall.

\noindent\textbf{Sample Dependencies.}
When sampling a tool $v$, an input parameter $p_i \in \mathcal{I}(v)$ is deemed \textit{independent} if it satisfies at least one of the following conditions:
1). \textit{Optional}: $p_i$ has a default value or can be omitted;
2). \textit{Externally providable}: $p_i$ is classified as \textit{external} so it can be naturally provided by the users;
3). \textit{Internally satisfiable}: $p_i$ is classified as \textit{internal} but it's also an output of previously sampled tool in $\tau$.
For any \textit{dependent} parameter $p_i$, the sampler recursively selects a \textit{prior} tool capable of generating it by traversing backward along the inverse edges of $G$. 
This recursive process ensures that all dependencies are resolved before $v$ is added to $\tau$. 
Additionally, to encourage diversity, the sampler may stochastically introduce a prior tool for a resolvable parameter with a small probability $p$. The full algorithmic details are provided in Appendix~\ref{appendix:algorithm}.

\noindent\textbf{Sample Neighbors.}  
Once all dependencies for $v$ are resolved, the sampler randomly selects $1$ to $k$ neighbors (with equal probability) along the outgoing edges from $v$ to extend the tool chain. 
This branching mechanism enables non-linear tool-use patterns beyond simple sequential chains, guiding more complex tool-use trajectory synthesis.
\subsection{Tool-Use Trajectory Synthesis}
\label{sec:query_gen}
\noindent\textbf{Overview.} Using a topology-aware sampling strategy, we sample tool chains $\tau$ subject to logical dependency constraints. Based on $\tau$, \textit{QueryGen} synthesizes multi-turn, multi-step tool-use trajectories through two principles: (1) \textit{Realistic user intent:} iteratively generating and refining naturalistic intents to reflect real-world pragmatic patterns such as implicit reasoning and ambiguity; and (2) \textit{Verifiable ground-truth:} deploying sandboxed agentic interaction to produce verified tool-call trajectories that ensure reliable reward signals. The prompts can be found in Appendix~\ref{appendix:prompts}.

\noindent\textbf{Planning.}
Grounded on $\tau$, we first construct a user profile and scenario. From this scenario, we derive a database state strictly conforming to the schema in Section~\ref{sec:database_modeling}. We then stochastically partition the tool chain into multiple dialogue turns, each comprising 1--5 randomly sampled tools.

\noindent\textbf{Generation and Refinement.}
For each turn, the \textit{QueryGen} synthesizes a naturalistic user query conditioned on the current database state, dialogue history, and sampled tools through two stages: (i) \textit{Subgoal decomposition}, where tools are broken into fine-grained subgoals and user intents, and (ii) \textit{Goal articulation}, where natural language requests are composed from these subgoals. Because initially generated queries often lack the implicit reasoning and conciseness characteristic of human language, the \textit{QueryGen} enhances realism through four calibrated refinement: (1) \textit{Implicit reference:} replacing explicit identifiers with contextual references and omitting deducible parameters; (2) \textit{Action compression:} compressing logically inferable intermediate steps; (3) \textit{Ambiguity introduction}: introducing reasonable referential ambiguity; and (4) \textit{Goal expansion:} augmenting queries with plausible, thematically related secondary objectives. With decomposition and refinement, the synthesized query reflects the pragmatic and implicit nature of real user requests.

\noindent\textbf{Agentic Interaction.}
To obtain ground-truth tool-call trajectories, we deploy sandbox environments with agents and simulated users, mirroring the RL training setup. At each turn, the agent resolves the generated queries by invoking tools, issuing explicit instructions to the user, or requesting clarification while the user follows the instructions from the agent, answers the questions, or proactively ends the conversation based on the feedback. This process continues until the user actively terminates the conversation or the maximum step limit is reached. We independently generate $k$ candidate solution trajectories to ensure comprehensive coverage of the solution space. Simulated users details can be found in Appendix~\ref{appendix:user}.

\noindent\textbf{Evaluation.}
Given $k$ candidate trajectories and their corresponding database state changes, the pipeline evaluates each solution and selects the one that optimally resolves the query. A filtering process then removes redundant tool calls and unnecessary user interactions, and a masking process annotates the arguments whose values do not affect tool-use correctness for each retained tools.
\subsection{Model Training}
\label{sec:reward}
With the synthesized trajectories, we perform post-training using both SFT and RL. For RL, evaluating tool-use correctness is non-trivial because valid executions are often non-unique and cannot be determined solely from reference trajectories or final database states. For example, independent read-only tool calls may be invoked in different orders, and parameters such as \texttt{limit} may vary across equally valid executions.
To account for this ambiguity, we use a composite reward with three components: 
1) \textit{trajectory-based reward} that measures the matches between the predicted and ground-truth tool-calling sequences; 
2) \textit{state-based reward} that evaluates the equivalence of the final database states after tool execution; and 
3) \textit{length penalty} that discourages unnecessarily long tool-call sequences. The overall reward is:
{
\setlength{\abovedisplayskip}{4pt}
\setlength{\belowdisplayskip}{4pt}
\[
R = \alpha \cdot R_{\text{traj}} + (1-\alpha) \cdot R_{\text{state}} - \gamma \cdot P_{\text{length}}
\]
}
where $R_{\text{traj}} \in [0, 1]$ is the trajectory-based reward, $R_{\text{state}}$ is the state-based reward, $P_{\text{length}}$ is the length penalty and $\alpha,\gamma\geq0$ are the weighting coefficients.
\section{Experiments and Analysis}
\label{experiments}
\subsection{Setup}
\label{sec:experiment-setup}

\noindent\textbf{Data Statistics.}
We construct 85 diverse MCP environments spanning seven domains: commerce, finance, travel, office, lifestyle, research, and utilities. From these environments, we synthesize 1,622 conversations for SFT trajectories and 953 conversations for RL trajectories. On average, each conversation comprises 4.82 turns, with each turn containing 3.29 steps—including both tool calls and user interactions. Further details are provided in Figure~\ref{fig:data_statistic}.

\noindent\textbf{Baselines and Benchmarks.}
We adopt Qwen3-(1.7B, 4B, 8B)~\citep{qwen3technicalreport2025} as training backbones. For baseline comparison, we directly use available checkpoints from AWM~\citep{awm2026} and EnvScaler~\citep{envscaler2026}, two concurrent work on tool-use trajectory synthesis. Evaluation is conducted on BFCL v3~\citep{bfcl2025}, $\tau^2$-Bench~\citep{tau2bench2025}, VitaBench~\citep{vitabench2025}, and MCP-Atlas~\citep{mcpatlas2026}.
Further details are provided in Appendix~\ref{appendix:implementation_details}.

\noindent\textbf{Implementation Details.}
Our training pipeline consists of: \textit{Stage 1}: SFT initialized with user interaction trajectories; \textit{Stage 2}: RL training uses only tool-call trajectories. We perform SFT using LlamaFactory~\citep{llamafactory2024} and RL using VeRL~\citep{verl2024} with GRPO~\citep{deepseekmath2024}. 
Details are provided in Appendix~\ref{appendix:implementation_details}.

\subsection{Main Results}
\begin{table*}[t]
\caption{\small Experiment results on BFCL, $\tau^2$-Bench, VitaBench, and MCP-Atlas. \colorbox{blue!15}{\textbf{Cell}} and \colorbox{blue!5}{\underline{Cell}} indicate the best and second-best results for each evaluation metric, respectively, while \colorbox{green!15}{\textbf{Cell}} and \colorbox{green!5}{\underline{Cell}} denote methods that achieve stronger performance with fewer environments and training tasks.}
\label{tab:experiment_all}
\centering
\begingroup
\setlength{\tabcolsep}{2.5pt}
\renewcommand{\arraystretch}{1.05}
\resizebox{\textwidth}{!}{%
\small
\begin{tabular}{@{}l c c c c c c c c c c c c c c c@{}}
\toprule
& \multicolumn{2}{c}{\textbf{Data Scale}}
& \multicolumn{2}{c}{\textbf{BFCL}}
& \multicolumn{2}{c}{\textbf{MCP-Atlas}}
& \multicolumn{4}{c}{\textbf{$\tau^2$-Bench}}
& \multicolumn{4}{c}{\textbf{VitaBench}}
& \textbf{Overall} \\
\cmidrule(lr){2-3}
\cmidrule(lr){4-5}
\cmidrule(lr){6-7}
\cmidrule(lr){8-11}
\cmidrule(lr){12-15}
\cmidrule(lr){16-16}
\textbf{Model}
& \textit{Env.}
& \textit{Tasks}
& \textit{\makecell{Single\\Turn}}
& \textit{\makecell{Multi\\Turn}}
& \textit{\makecell{Pass\\Rate}}
& \textit{\makecell{Mean\\Cov.}}
& \textit{Airline}
& \textit{Retail}
& \textit{Tele}
& \textbf{Avg.}
& \textit{Deliver}
& \textit{Store}
& \textit{Ota}
& \textbf{Avg.}
& \textit{Avg.} \\
\midrule

\rowcolor{gray!10} \multicolumn{16}{c}{\textbf{Qwen3-1.7B}} \\
Base 
& -- & -- 
& \colorbox{blue!15}{\textbf{79.48}} & 16.75 
& 1.03 & 6.25 
& 14.00 & 7.02 & 22.81 & 14.61 
& 4.00 & 0.00 & 0.00 & 1.33 
& 16.27 \\

EnvScaler 
& 191 & 11,572 
& 60.41 & \colorbox{blue!15}{\textbf{30.13}} 
& \colorbox{blue!5}{\underline{2.75}} & 9.40 
& 12.00 & 18.42 & 10.53 & 13.65 
& 9.00 & 3.00 & 1.09 & 4.36 
& 16.51 \\

Our (SFT) 
& \colorbox{green!15}{\textbf{85}} & \colorbox{green!15}{\textbf{1,622}} 
& 78.30 & 23.25 
& 1.72 & \colorbox{blue!15}{\textbf{10.05}} 
& 16.00 & 20.18 & 10.53 & \colorbox{blue!15}{\textbf{15.57}} 
& 13.00 & 6.00 & 0.00 & \colorbox{blue!5}{\underline{6.33}} 
& \colorbox{blue!5}{\underline{18.60}} \\

Our 
& \colorbox{green!15}{\textbf{85}} & \colorbox{green!5}{\underline{2,575}} 
& \colorbox{blue!5}{\underline{78.44}} & \colorbox{blue!5}{\underline{28.38}} 
& \colorbox{blue!15}{\textbf{3.09}} & \colorbox{blue!5}{\underline{9.64}} 
& 12.00 & 16.67 & 16.67 & \colorbox{blue!5}{\underline{15.11}} 
& 11.00 & 11.00 & 0.00 & \colorbox{blue!15}{\textbf{7.33}} 
& \colorbox{blue!15}{\textbf{19.74}} \\

\midrule

\rowcolor{gray!10} \multicolumn{16}{c}{\textbf{Qwen3-4B}} \\
Base 
& -- & -- 
& 85.15 & 33.50 
& 4.12 & 12.86 
& 24.00 & 38.60 & 13.16 & 25.25 
& 9.00 & 12.00 & 2.02 & 7.67 
& 24.09 \\

AWM 
& 526 & 3,315 
& \colorbox{blue!15}{\textbf{85.97}} & 40.75 
& 4.47 & 12.33 
& 18.00 & 31.58 & 17.54 & 22.37 
& 22.00 & 13.00 & 0.00 & 11.67 
& 25.47 \\

EnvScaler 
& 191 & 11,572 
& 83.64 & \colorbox{blue!5}{\underline{45.00}} 
& \colorbox{blue!15}{\textbf{9.97}} & \colorbox{blue!15}{\textbf{22.27}} 
& 36.00 & 41.23 & 10.53 & \colorbox{blue!5}{\underline{29.25}} 
& 23.00 & 15.00 & 6.06 & \colorbox{blue!5}{\underline{14.69}} 
& \colorbox{blue!5}{\underline{29.56}} \\

Our (SFT) 
& \colorbox{green!15}{\textbf{85}} & \colorbox{green!15}{\textbf{1,622}} 
& 85.10 & 44.25 
& 7.90 & 19.66 
& 24.00 & 47.37 & 4.39 & 25.25 
& 19.00 & 12.00 & 3.00 & 11.33 
& 27.29 \\

Our 
& \colorbox{green!15}{\textbf{85}} & \colorbox{green!5}{\underline{2,575}} 
& \colorbox{blue!5}{\underline{85.46}} & \colorbox{blue!15}{\textbf{48.50}} 
& \colorbox{blue!15}{\textbf{9.97}} & \colorbox{blue!5}{\underline{21.89}} 
& 36.00 & 42.11 & 12.28 & \colorbox{blue!15}{\textbf{30.13}} 
& 21.00 & 21.00 & 6.00 & \colorbox{blue!15}{\textbf{16.00}} 
& \colorbox{blue!15}{\textbf{30.77}} \\

\midrule

\rowcolor{gray!10} \multicolumn{16}{c}{\textbf{Qwen3-8B}} \\
Base 
& -- & -- 
& 84.31 & 41.25 
& 5.15 & 14.86 
& 32.00 & 42.98 & 21.93 & 32.30 
& 24.00 & 15.00 & 11.11 & 16.70 
& 29.23 \\

AWM 
& 526 & 3,315 
& 84.80 & 42.25 
& 6.19 & 16.60 
& 30.00 & 29.82 & 25.44 & 28.42 
& 20.00 & 15.00 & 14.43 & 16.48 
& 28.65 \\

EnvScaler 
& 191 & 11,572 
& 84.74 & \colorbox{blue!15}{\textbf{51.88}} 
& \colorbox{blue!5}{\underline{9.62}} & 22.63 
& 38.00 & 49.12 & 15.79 & \colorbox{blue!15}{\textbf{34.30}} 
& 25.00 & 19.00 & 12.00 & \colorbox{blue!15}{\textbf{18.67}} 
& \colorbox{blue!5}{\underline{32.72}} \\

Our (SFT) 
& \colorbox{green!15}{\textbf{85}} & \colorbox{green!15}{\textbf{1,622}} 
& \colorbox{blue!5}{\underline{84.83}} & 46.50 
& 8.25 & \colorbox{blue!5}{\underline{22.86}} 
& 42.00 & 43.86 & 12.28 & 32.71 
& 23.00 & 20.00 & 7.00 & 16.67 
& 30.82 \\

Our 
& \colorbox{green!15}{\textbf{85}} & \colorbox{green!5}{\underline{2,575}} 
& \colorbox{blue!15}{\textbf{86.02}} & \colorbox{blue!5}{\underline{49.00}} 
& \colorbox{blue!15}{\textbf{13.75}} & \colorbox{blue!15}{\textbf{25.98}} 
& 44.00 & 43.86 & 13.16 & \colorbox{blue!5}{\underline{33.67}} 
& 24.00 & 22.00 & 10.00 & \colorbox{blue!15}{\textbf{18.67}} 
& \colorbox{blue!15}{\textbf{33.40}} \\

\bottomrule
\end{tabular}%
}
\endgroup
\end{table*}

Table~\ref{tab:experiment_all} presents a comprehensive comparison across four benchmarks and strong baselines. 

\noindent\textbf{SFT Cold Start Delivers the Largest Relative Gains.}
Supervised fine-tuning on our automatically generated trajectories alone yields substantial improvements across diverse tool-use benchmarks. On BFCL multi-turn evaluation, \our (SFT) improves Qwen3-1.7B from 16.75 to 23.25 and Qwen3-4B from 33.50 to 44.25. Similar gains are observed on $\tau^2$-Bench, where Qwen3-1.7B improves from 14.61 to 15.57, while Qwen3-4B achieves a strong gain on the challenging retail domain (38.60 $\rightarrow$ 47.37). The improvements further generalize to more challenging benchmarks. On MCP-Atlas, pass rates nearly double across all model scales, e.g., from 4.12 to 7.90 for Qwen3-4B and from 5.15 to 8.25 for Qwen3-8B. On VitaBench, Qwen3-1.7B improves from 1.33 to 6.33, while Qwen3-4B improves from 7.67 to 11.33. Overall, \our (SFT) consistently improves average performance across all model scales, demonstrating that our synthesized trajectories provide an effective cold-start signal for scalable tool-use learning.

\noindent\textbf{RL after SFT Further Unlocks Tool-Use Capability.}
Building on the strong SFT initialization, RL training consistently yields further gains across nearly all benchmarks and model scales.
Compared with \our (SFT), the full \our improves the overall score from 18.60 to 19.74 for Qwen3-1.7B, from 27.29 to 30.77 for Qwen3-4B, and from 30.82 to 33.40 for Qwen3-8B. The improvements are particularly evident on challenging interactive benchmarks. On VitaBench, Qwen3-4B improves from 11.33 to 16.00, while on MCP-Atlas, Qwen3-8B substantially improves pass rate from 8.25 to 13.75 and mean coverage from 22.86 to 25.98. Similar gains are observed on BFCL multi-turn evaluation, where Qwen3-4B improves from 44.25 to 48.50 and Qwen3-8B from 46.50 to 49.00. These results suggest that SFT provides strong foundational tool-use behaviors and RL further enhances 
reasoning and execution robustness.

\noindent\textbf{Strong Generalization Across Benchmark Types.}
\our demonstrates consistent improvements across both conversational benchmarks ($\tau^2$-Bench and VitaBench) and non-conversational benchmarks (BFCL and MCP-Atlas). On conversational benchmarks, Qwen3-4B improves from 25.25 to 30.13 on $\tau^2$-Bench and from 7.67 to 16.00 on VitaBench, while Qwen3-8B achieves the best conversational performance with 33.67 and 18.67, respectively. At the same time, \our substantially improves non-conversational tool-use capability, boosting BFCL multi-turn accuracy from 33.50 to 48.50 for Qwen3-4B and achieving the best MCP-Atlas results with a 13.75 pass rate and 25.98 mean coverage on Qwen3-8B. These results demonstrate that \our generalizes effectively across both conversational interaction and compositional tool-execution settings.
\subsection{Effect of the Environments Scaling}
\begin{figure*}[h]
    \centering
    \includegraphics[width=0.95\textwidth]{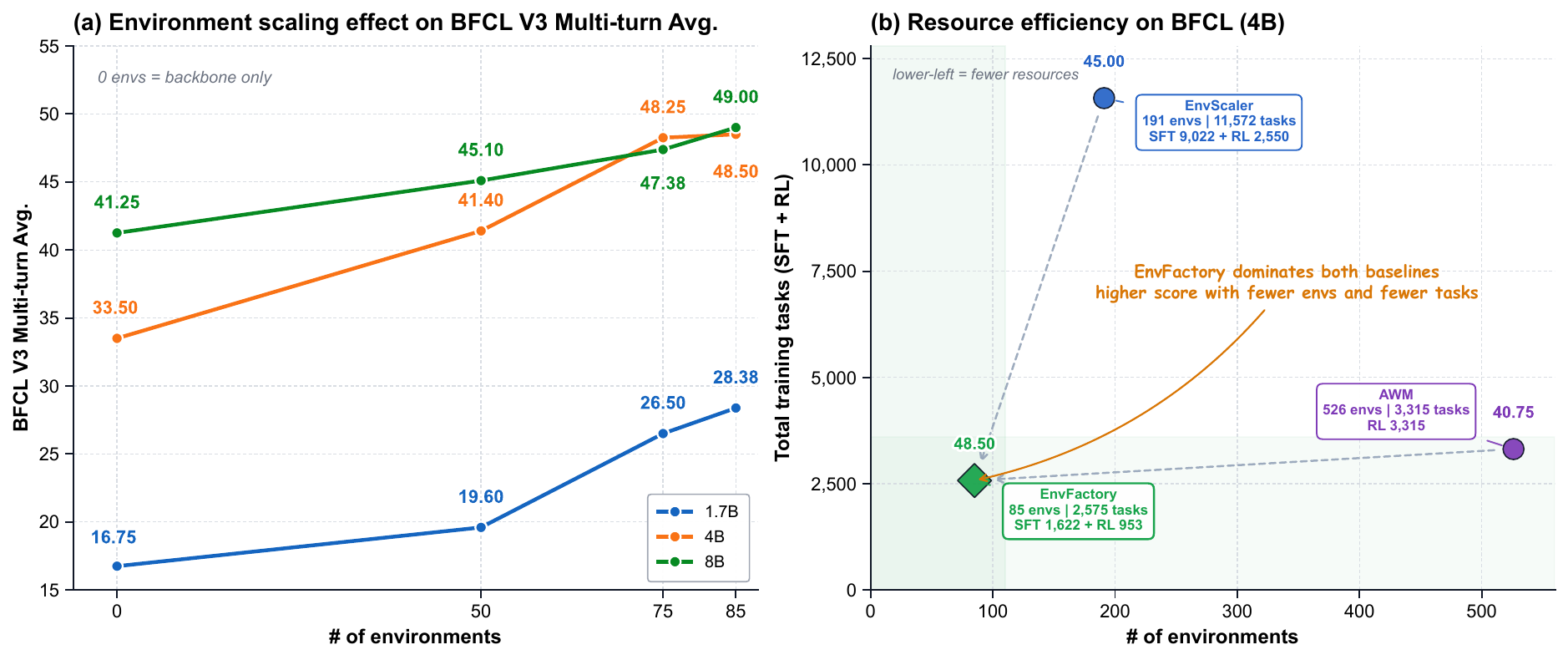}
    \caption{\small Environment scaling and resource efficiency analysis on BFCL-v3. (a) BFCL-v3 multi-turn average performance under different numbers of environments across Qwen3-1.7B, Qwen3-4B, and Qwen3-8B. (b) Resource efficiency comparison on Qwen3-4B, where the x-axis denotes the number of environments, the y-axis denotes the total number of training tasks, and the marker label reports the BFCL-v3 multi-turn average score.}
    \label{fig:env_scaling}
\end{figure*}

Figure~\ref{fig:env_scaling} studies how the number of executable environments affects tool-use learning in \our. To evaluate scaling behavior, we construct two additional training subsets with 50 and 75 randomly sampled environments, respectively, and perform the same SFT+RL training procedure on each subset. As shown in Figure~\ref{fig:env_scaling}(a), increasing the environment pool consistently improves BFCL-v3 multi-turn performance across Qwen3-1.7B, Qwen3-4B, and Qwen3-8B. This trend indicates that broader environment coverage exposes the model to more diverse tool schemas, state transitions, and multi-step interaction patterns, improving generalization to unseen tool-use tasks. The scaling curve also shows a diminishing-return pattern: the gain from 50 to 75 environments is larger than that from 75 to 85 environments, suggesting that later additions may contain more overlapping tool logic or task structures. Figure~\ref{fig:env_scaling}(b) further shows that \our achieves stronger BFCL-v3 multi-turn performance while using only 85 environments and 2,575 training tasks, far fewer than the baselines. This result suggests that verified stateful environments and dependency-aware trajectories provide effective supervision and reward signals from a compact training set.

\subsection{Ablation Study}
\begin{table}[H]
\centering
\begin{minipage}[t]{0.485\textwidth}
\vspace{0pt}
\centering
\caption{\small Performance of training with direct RL.}
\label{tab:ablation_rl}
\vspace{0.78\baselineskip}
\small
\setlength{\tabcolsep}{2.4pt}
\renewcommand{\arraystretch}{1.16}
\resizebox{\linewidth}{!}{%
\begin{tabular}{@{}l c c c c@{}}
\toprule
\textbf{Model} &
\textbf{\makecell{BFCL\\Single-turn}} &
\textbf{\makecell{BFCL\\Multi-turn}} &
\textbf{\makecell{$\tau^2$\\Bench}} &
\textbf{\makecell{VitaBench}} \\
\midrule
Qwen3-1.7B & 79.48 & 16.75 & 14.67 & 1.33 \\
Our-1.7B (RL) & \cellcolor{blue!10}\textbf{79.53} & \cellcolor{blue!10}\textbf{18.33} & \cellcolor{blue!10}\textbf{18.28} & \cellcolor{blue!10}\textbf{1.67} \\
\midrule
Qwen3-4B & 85.15 & 33.50 & \cellcolor{blue!10}\textbf{25.33} & 7.67 \\
Our-4B (RL) & \cellcolor{blue!10}\textbf{85.26} & \cellcolor{blue!10}\textbf{41.38} & 24.83 & \cellcolor{blue!10}\textbf{12.74} \\
\midrule
Qwen3-8B & 84.31 & 41.25 & \cellcolor{blue!10}\textbf{32.33} & 16.70 \\
Our-8B (RL) & \cellcolor{blue!10}\textbf{84.42} & \cellcolor{blue!10}\textbf{44.35} & 29.08 & \cellcolor{blue!10}\textbf{17.00} \\
\bottomrule
\end{tabular}%
}
\end{minipage}
\hfill%
\begin{minipage}[t]{0.485\textwidth}
\vspace{-0.12\baselineskip}
\centering
\caption{\small Performance comparison between refined and unrefined trajectories for SFT.}
\label{tab:refinement_effect}
\small
\setlength{\tabcolsep}{2.0pt}
\resizebox{\linewidth}{!}{%
\begin{tabular}{@{}l c c c c c@{}}
\toprule
\textbf{Model} &
\textbf{Base} &
\textbf{\makecell{Miss\\Func}} &
\textbf{\makecell{Miss\\Param}} &
\textbf{\makecell{Long\\Context}} &
\textbf{Overall} \\
\midrule
Unrefine-1.7B & 30.0 & 21.5 & 19.5 & 14.0 & 21.25 \\
Refine-1.7B & \cellcolor{blue!10}\textbf{30.5} & \cellcolor{blue!10}\textbf{22.5} & \cellcolor{blue!10}\textbf{21.0} & \cellcolor{blue!10}\textbf{14.5} & \cellcolor{blue!10}\textbf{22.12} \\
\midrule
Unrefine-4B & \cellcolor{blue!10}\textbf{52.0} & 47.0 & 30.5 & 34.0 & 40.88 \\
Refine-4B & 49.5 & \cellcolor{blue!10}\textbf{47.5} & \cellcolor{blue!10}\textbf{32.0} & \cellcolor{blue!10}\textbf{36.0} & \cellcolor{blue!10}\textbf{41.25} \\
\midrule
Unrefine-8B & 51.5 & 47.0 & 38.5 & 35.0 & 43.00 \\
Refine-8B & \cellcolor{blue!10}\textbf{55.0} & 47.0 & \cellcolor{blue!10}\textbf{39.0} & 35.0 & \cellcolor{blue!10}\textbf{44.00} \\
\bottomrule
\end{tabular}%
}
\end{minipage}

\end{table}

\noindent\textbf{Experiment Results on Direct RL.} \par
\noindent We examine whether \our-generated trajectories can directly support RL training without an SFT cold-start phase. As shown in Table~\ref{tab:ablation_rl}, direct RL improves several interactive benchmarks, such as BFCL multi-turn accuracy for \our-4B (33.50 to 41.38) and $\tau^2$-Bench for \our-1.7B (14.67 to 18.28).
However, these gains are smaller and less stable than RL after SFT, indicating that SFT initialization remains important for stable policy optimization.

\noindent\textbf{Effects of the Refinement Stage.} 
To study the impact of the refinement stage in query generation, we synthesize 250 SFT trajectories with and without refinement, respectively. Table~\ref{tab:refinement_effect} shows that refined trajectories consistently outperform unrefined ones, especially on ambiguous settings such as \texttt{Miss-Func} and \texttt{Miss-Param}. This suggests that refinement improves query ambiguity calibration and provides higher-quality supervision.

\begin{wrapfigure}{r}{0.60\textwidth}
\vspace{-1.0em}
    \centering
    \includegraphics[width=\linewidth]{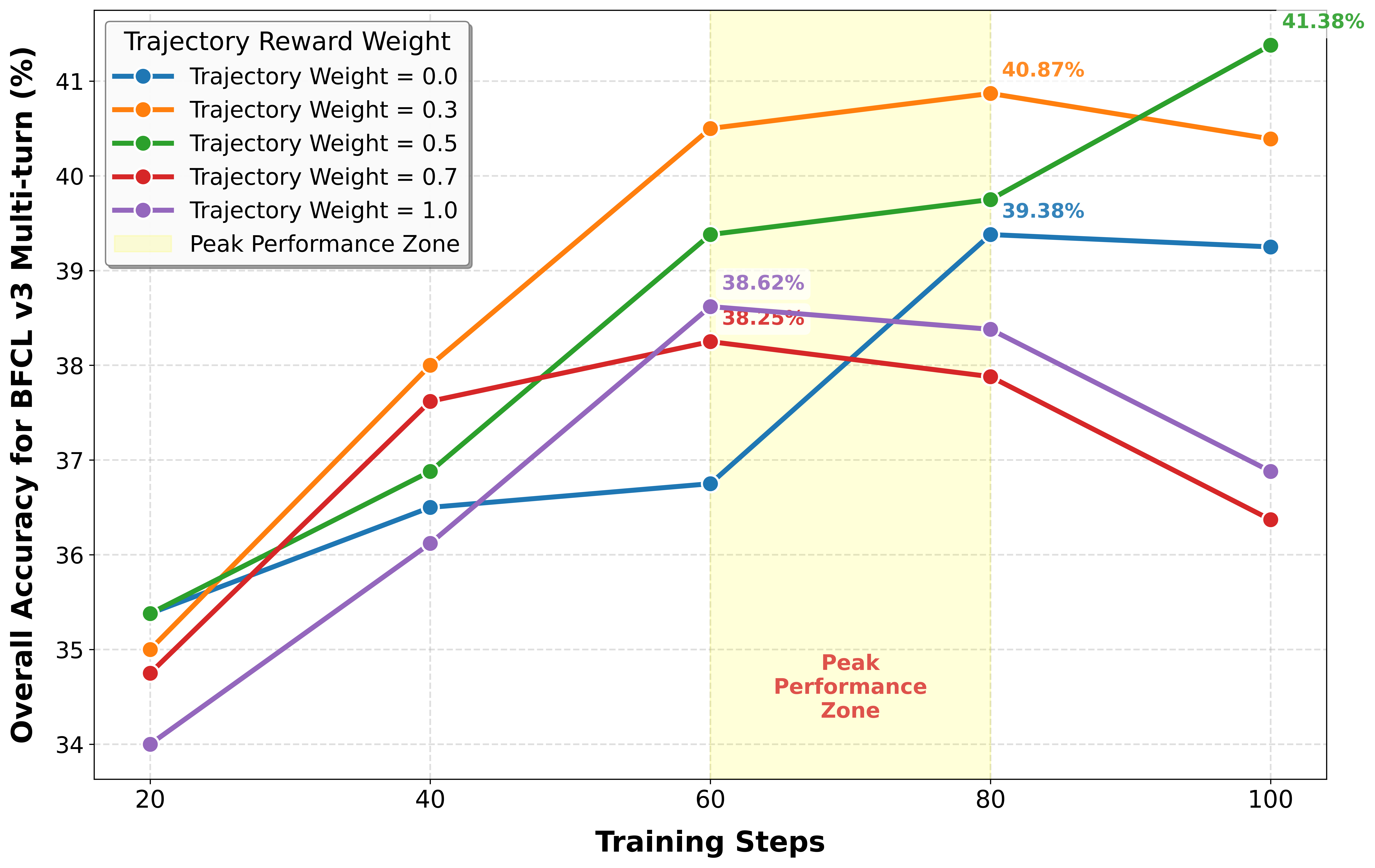}
    \vspace{-2mm}
    \caption{\small Ablation results on BFCL-v3 under different trajectory reward weights.}
    \label{fig:reward_weighting}
\vspace{-1.0em}
\end{wrapfigure}
\noindent\textbf{Effects of the Reward Weighting Coefficient.} 
We conduct an ablation over the trajectory-based reward weighting coefficient $\alpha \in \{0, 0.3, 0.5, 0.7, 1.0\}$ on BFCL while fixing the length penalty coefficient $\gamma$. Figure~\ref{fig:reward_weighting} shows that relying only on state-based reward ($\alpha=0$) or only on trajectory matching ($\alpha=1.0$) degrades performance. Balanced weighting performs better, with $\alpha=0.5$ achieving the best peak accuracy of $41.38\%$. Removing either reward component altogether hurts performance, indicating that both trajectory fidelity and state equivalence are necessary for effective RL training.


\section{Conclusion}
We presented \our, a fully automated framework that addresses two critical bottlenecks in Agentic RL for tool-use: the lack of scalable, verifiable environments and the scarcity of realistic, implicitly-reasoned training trajectories. Unlike prior approaches that rely on costly production APIs, hallucination-prone simulators, or static synthetic environments, \our autonomously constructs verified, stateful environments by exploring real-world online resources and recursively resolving logical dependencies among tools. It further bridges the realism gap by transforming over-specified instruction lists into natural human-like requests through calibrated refinement that injects implicit intents and ambiguity. 
Experimental results show that \our consistently outperforms strong baselines in both training efficiency and downstream performance, while requiring significantly fewer synthetic environments and samples. 
%

\clearpage

\bibliography{main}
\bibliographystyle{abbrvnat}

\clearpage

\appendix
\section{Broader Impact}
\label{appendix:impact}
This work introduces a framework for the automated construction of executable environments and realistic trajectories, significantly lowering the barrier for developing robust AI agents capable of complex tool-use. By providing a scalable alternative to costly production APIs and hallucination-prone simulations, our approach facilitates the democratization of Agentic RL research, enabling a broader range of researchers to train agents in diverse, high-fidelity domains such as finance, research, and office automation. Furthermore, by injecting realistic human communication patterns—such as implicit intents and ambiguity—into synthetic data, this research moves AI agents closer to safe and effective real-world deployment, ensuring they can better interpret and act upon human needs.

However, the automation of agent training and environment synthesis carries potential risks that necessitate responsible oversight. The ability to rapidly generate executable tool-use ecosystems could be misused to simulate and automate malicious activities, such as large-scale fraudulent financial transactions or sophisticated phishing campaigns, if applied to sensitive domains without safeguards. Additionally, since the framework relies on online resources and LLM-guided proposals, it may inadvertently encode or amplify biases present in its source data or underlying models. To mitigate these risks, we have documented our dataset and environment construction process transparently, released our artifacts under restrictive licenses to prevent misuse, and encourage the integration of rigorous safety constraints within the synthesized environments to ensure that agents remain aligned with ethical and legal standards.

\section{Limitations}
\label{sec:limitations}
\our uses the MCP as its tool interface. The MCP servers we design are stateful: write‑capable tools can modify a shared environment database, which forces strict session isolation to prevent cross‑contamination. As a result, each conversation requires a dedicated transport connection to the target servers, constraining the degree of parallel tool invocation and creating a throughput bottleneck during large‑scale data synthesis. We mitigate this limitation by implementing an asynchronous synthesis pipeline that executes many isolated sessions concurrently, thereby maximizing overall generation efficiency despite the per‑connection requirement.


\section{LLM Usage Declaration}
\label{appendix:llmusage}
This manuscript uses LLMs strictly for the purpose of language editing and textual polishing to enhance presentation quality. We declare that the novel ideas, methodological framework, experimental execution, and data analysis are the original work of the authors. All content modified by AI tools has been carefully reviewed and validated by the authors to ensure accuracy.

\section{Compute Usage}
\label{appendix:compute_usage}
\noindent\textbf{GPU Usage.}
We report the GPU resources required for each stage of our pipeline. 
For SFT data synthesis, we deploy \texttt{Qwen3-30B-A3B-Thinking-2507} on $2\times 80$GB GPUs to generate data and distill reasoning processes. Synthesizing 1,000 multi-turn, multi-step trajectories requires approximately $20$ GPU hours.

For SFT training, we fine-tune \texttt{Qwen3-4B} for $3$ epochs using LlamaFactory~\citep{llamafactory2024} on $8\times 80$GB GPUs, which consumes around $10$ GPU hours.

For RL training, we train \texttt{Qwen3-4B} for $10$ epochs using VeRL~\citep{verl2024} on $8\times 80$GB GPUs, requiring approximately $20$ GPU hours.

\noindent\textbf{Token Usage.}
\label{appendix:token_usage}
\our can autonomously scale up environments and data generation. The table below summarizes our token consumption across different stages. We note that trajectory synthesis supports asynchronous generation, enabling efficient scaling: synthesizing 1,000 multi-turn, multi-step trajectories takes roughly $20$ hours (approximately $1.2$ minutes per conversation).
\begin{table}[h]
\centering
\resizebox{\textwidth}{!}{%
\begin{tabular}{lcccrc}
\hline
\textbf{Mode} & \textbf{Model} & \textbf{Prompt Token} & \textbf{Completion Token} & \textbf{GPU Time} & \textbf{Success Rate} \\
\hline
Environment & \texttt{Kimi-K2-Thinking} & 192K & 31K & 3 min & 92.9\% \\
SFT Trajectory & \texttt{Qwen3-30B-A3B-Thinking} & 228K & 84K & 6 min & 85.4\% \\
RL Trajectory & \texttt{DeepSeek-V3.2} & 195K & 19K & 3 min & 88.2\% \\
\hline
\end{tabular}%
}
\caption{Token consumption across environment construction and query synthesis.}
\label{tab:token_consumption}
\end{table}

\section{Additional Related Work}
\label{appendix:related}
\paragraph{Reinforcement Learning for LLMs.} Reinforcement Learning (RL) has become a cornerstone of LLM post-training. Following the early adoption of reward-model-based pipelines \citep{ouyang2022training}, Direct Preference Optimization \citep{Rafailov2023} streamlined this process by directly leveraging pairwise preference data. More recently, Reinforcement Learning with Verifiable Rewards (RLVR) has significantly pushed the boundaries of downstream performance in mathematics, coding, and agentic tasks. A prominent example is GRPO \citep{shao2024deepseekmath}, which optimizes LLMs at the group level by aggregating multiple outputs to provide diverse preference signals, thereby improving generalization. To achieve more fine-grained optimization, TreeRPO \citep{yang2025treerpo} extends GRPO by replacing sparse, trajectory-level rewards with tree-sampled, step-level dense rewards to better guide intermediate reasoning steps.

Despite these advancements, the fundamental mechanics of RLVR remain under scrutiny. Notably, \citet{yue2025does} questioned whether RLVR truly expands a base model's intrinsic capabilities, demonstrating through experiments that it fails to improve Pass@k—a metric tightly coupled with an LLM's reasoning upper bound. This limitation is often attributed to a rapid decline in model output entropy during the early stages of RLVR training, which stifles sustained exploration later on \citep{gao2025one,zhu2025surprising}. To mitigate this exploration collapse, SvS \citep{liang2025beyond} introduces a self-play-style problem augmentation strategy that enhances training data diversity, successfully stabilizing entropy and significantly boosting Pass@k performance. Alternatively, DARS \citep{yang2025depth} addresses these training biases through difficulty-adaptive rollout sampling combined with large-batch training, ultimately delivering robust improvements in both Pass@1 and Pass@k reasoning performance.

\section{Implementation Details}
\label{appendix:implementation_details}

\noindent\textbf{Data Synthesis Setup.} In the \textbf{EnvGen} pipeline of \our, we primarily leverage Kimi-K2-Thinking~\citep{kimik2technicalreport2025} to propose, draft, construct, and verify MCP environments. For the \textbf{QueryGen} pipeline, we employ DeepSeek-V3.2-Chat~\citep{deepseekv32technicalreport2025} for RL tool-use trajectories generation, while utilizing Qwen3-30B-A3B-Thinking-2507~\citep{qwen3technicalreport2025} SFT tool-use trajectories synthesis to distill the thinking process for SFT.

\noindent\textbf{Reinforcement Learning Setup.} 
We employ Group Relative Policy Optimization (GRPO)~\citep{deepseekmath2024} implemented with the Verl framework~\citep{verl2024}. Training is conducted on $8 \times 80\,\text{GB}$ GPUs using a learning rate of $1 \times 10^{-6}$, rollout size of $8$, and batch size of $256$. We set the maximum trajectory length to $16$k tokens and the maximum generation length to $4$k tokens, and train for $10$ epochs. For RL training, each interaction turn is treated as an individual training sample.

\noindent\textbf{Supervised Fine-Tuning Setup.} 
We perform SFT using LlamaFactory~\citep{llamafactory2024} on $8 \times 80\,\text{GB}$ GPUs with a learning rate of $1 \times 10^{-6}$ and batch size of $256$, training for $3$ epochs. For subsequent RL training, we initialize from the checkpoint saved after the first SFT epoch. During SFT data construction, each tool-call or user-interaction step is treated as a separate training sample together with its associated reasoning trace. Failed tool calls are filtered out from the training data.

\noindent\textbf{Evaluation Setup.} 
During inference, we leverage the SGLang framework~\citep{sglang2024}. 
We set the sampling temperature to $0$ for non-thinking models and $0.7$ for thinking models, with tensor parallelism (TP) set to $2$ by default. 
For the user and evaluator agents in $\tau^2$-Bench and VitaBench, we employ DeepSeek-V3.2-Chat~\citep{deepseekv32technicalreport2025}.

\noindent\textbf{MCP-Atlas Setup.}
Due to network connectivity constraints, our evaluation of MCP-Atlas uses a subset comprising 30 of 36 servers and 291 of 500 tasks. The following servers are excluded: \texttt{mongodb}, \texttt{oxylabs}, \texttt{brave-search}, \texttt{wikipedia}, \texttt{slack}, and \texttt{google-workspace}.

\noindent\textbf{Simulated User Details.}
\label{appendix:user}
To instantiate a faithful simulation of tool-use scenarios, we first classify available MCP tools into \textit{user tools} and \textit{assistant tools} via LLM-based categorization. User tools comprise operations that are either: (i) \textit{confidential or sensitive} (e.g., \texttt{login}, \texttt{reset\_password}), or (ii) \textit{physically constrained} (e.g., \texttt{restart\_engine}). These tools require direct user authorization or physical presence and cannot be autonomously executed by the agent.

We then construct a simulated user by conditioning an LLM on three contextual inputs: (a) the narrative scenario, (b) the dialogue history, and (c) the current database state. To ensure realistic behavior, we constrain the user's knowledge to \textit{external parameters} identified in Section~\ref{sec:internal_external_param}---information that human users can realistically provide (e.g., personal preferences, location, time constraints). This prevents the simulated user from accessing \textit{internal parameters} (e.g., system-generated IDs, backend state) that would be unavailable to actual users, thereby avoiding implausible responses such as verbatim recitation of complex internal identifiers.

\section{Data Statistic}
\definecolor{green}{HTML}{2ECC71}

\newcommand{\greencheck}{%
  {\scalebox{0.75}{\textcolor{green}{\faCheckCircle}}}%
}

\setlength{\textfloatsep}{10pt}
\begin{table*}[h]
\caption{\small Comparison of environments and training samples between baselines with \greencheck\ indicates higher efficiency.}
\label{tab:comparison}
\centering
\resizebox{0.7\textwidth}{!}{%
\small
\begin{tabular}{@{}l c c c@{}}
\toprule
\textbf{Pipeline} & \textbf{Environments \#} & \textbf{SFT Tasks \#} & \textbf{RL Tasks \#} \\
\midrule
AWM~\citep{awm2026} & 526 & - & 3315 \\
EnvScaler~\citep{envscaler2026} & 191 & 9022 & 2550 \\
\our & 85\,\greencheck & 1622\,\greencheck & 953\,\greencheck \\
\bottomrule
\end{tabular}%
}
\end{table*}
\begin{figure*}[h]
    \centering
    \includegraphics[width=0.99\textwidth]{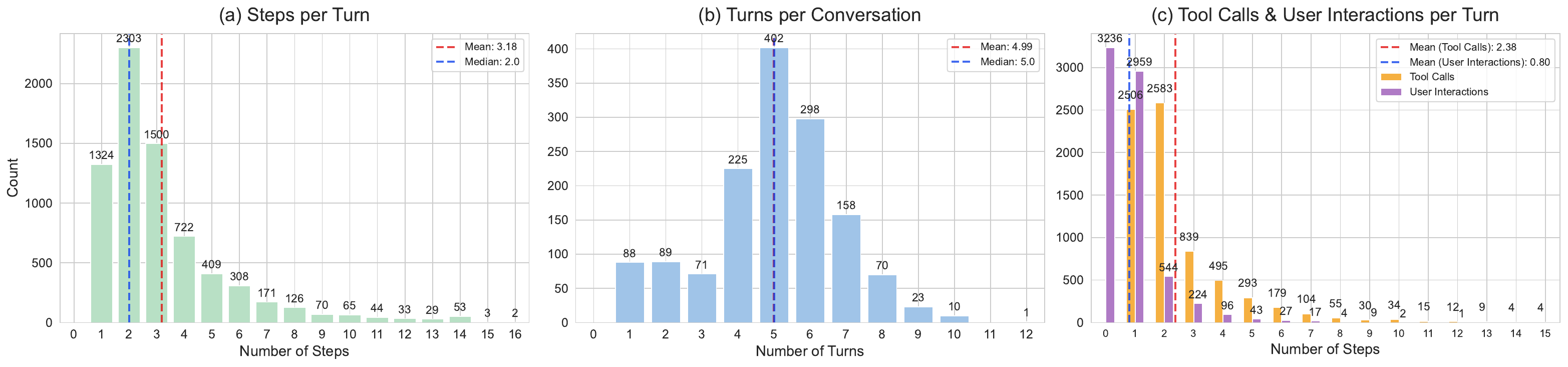}
    \caption{Distribution of conversation statistics. (a) Number of total steps per turn. (b) Number of turns per conversation. (c) Number of tool calls steps and user interactions per turn respectively.}
    \label{fig:data_statistic}
\end{figure*}

\section{Algorithms}
\label{appendix:algorithm}
Our topology-aware sampling strategy ensures execution feasibility by guaranteeing all required inputs $\mathcal{I}(v)$ of each sampled tool $v$ are satisfied before inclusion—addressing a key limitation of naive random walks. Operating on the directed dependency graph $G=(V,E)$ (Section~\ref{sec:topology_sampling}), the algorithm proceeds in two phases for each node $v$:

\noindent\textbf{Backward dependency resolution.}
Before adding $v$ to the visited set $\hat{V}$, the algorithm recursively resolves unsatisfied inputs via \texttt{SAMPLEPRIORS}. A parameter $p_i \in \mathcal{I}(v)$ is valid if: (1) optional (has schema default), (2) user-providable per LLM classification, or (3) already produced by some $u \in \hat{V}$ where $p_i \in \mathcal{O}(u)$. Invalid parameters trigger backward traversal to uniformly sample a producer tool $u$ satisfying $(u \rightarrow v) \in E$ and $p_i \in \mathcal{O}(u)$, with recursion depth capped at $D_{\max}=3$. A stochastic override ($p=0.1$) occasionally introduces additional priors for valid parameters to enhance trajectory diversity.

\noindent\textbf{Forward expansion.}
Once all dependencies are resolved and incorporated into $\hat{V}$, $v$ is added to $\hat{V}$ and the algorithm samples one outgoing neighbor from $N(v) = \{u \mid (v \rightarrow u) \in E\}$ for subsequent processing.
\begin{algorithm}[H]
\caption{Topology-based Sampling Strategy}
\label{alg:topology_sampling}
\begin{algorithmic}[1]
\Require $G = (V, E)$ with $|V| = N$, integer $n \leq N$, and start node $v_s$
\Ensure Sampled nodes $\hat V\subseteq V$ with $|\hat V|=n$
\State \textcolor{blue}{\textit{// Initialize visited nodes set and queue for BFS}}
\State $\hat V \gets \{v_s\}$ and $\text{queue} \gets \text{Queue}(v_s)$
\While{$|\hat V|<n$}
    \State $v\gets \text{queue}.\text{dequeue}()$
    \State \textcolor{blue}{\textit{// Sample priors for current node $v$}}
    \State $P(v)\gets\textsc{SamplePriors}(G,\hat V,v,0,D_{max})$
    \For{$p \in P(v)$}
        \If{$p \notin \hat V$}
            \State $\hat V\gets \hat V\cup\{p\}$
        \EndIf
    \EndFor
    \State $\hat V\gets \hat V\cup\{v\}$
    \State \textcolor{blue}{\textit{// Find all neighbors of $v$}}
    \State $N(v)\gets\{u\in V\mid e_{vu}=1, \forall e_{uv}\in E\}$
    \State \textcolor{blue}{\textit{// Randomly sample a neighbor of $v$}}
    \State $u\gets\text{Uniform}(N(v))$
    \State $\text{queue}.\text{enqueue}(u)$
\EndWhile
\State \textbf{return} $\hat V$
\end{algorithmic}
\end{algorithm}

\begin{algorithm}[H]
\caption{Sample Priors}
\label{alg:sample_priors}
\begin{algorithmic}[1]
\Require Graph $G = (\mathcal{V}, \mathcal{E})$, visited nodes $\hat{V}$, current node $v$, current depth $d$, max depth $D_{\max}$
\Ensure Set of sampled prior nodes $P_v$
\State $P_v \gets \emptyset$
\If{$d\geq D_{\max}$}
    \State \Return $P_v$
\EndIf
\For{each input parameter $p_i \in \mathcal{I}(v)$}
    \State \textcolor{blue}{\textit{// Skip if $p_i$ is valid unless stochastically overridden}}
    \If{$\textsc{IsValid}(p_i, \hat{V})$ and $\mathcal{U}(0,1)<0.1$}
        \State \textbf{continue}
    \EndIf
    \State \textcolor{blue}{\textit{// Find tools that output $p_i$}}
    \State $\mathcal{C} \gets \{ u \in \mathcal{V} \mid p_i \in \mathcal{O}(u) \text{ and } (u \rightarrow v) \in \mathcal{E} \}$
    \If{$\mathcal{C} = \emptyset$}
        \State \textbf{continue}
    \EndIf
    \State \textcolor{blue}{\textit{// Randomly select one prior tool}}
    \State $u \gets \text{Uniform}(\mathcal{C})$
    \If{$u \notin \hat{V}$ \textbf{and} $d < D_{\max}$}
        \State $P_u \gets \textsc{SamplePriors}(G, \hat{V}, u, d + 1, D_{\max})$
        \State $\hat{V} \gets \hat{V} \cup \{u\}$
        \State $P_v \gets P_v \cup \{u\} \cup P_u$
    \EndIf
\EndFor
\State \Return $P_v$
\end{algorithmic}
\end{algorithm}

\section{Prompts}
\label{appendix:prompts}
\subsection{Prompts for EnvGen}
\begin{tcolorbox}[
    breakable,
    colback=gray!20,
    colframe=gray!50,
    title=Tool Generation Prompt,
    fontupper=\small
]

\subsection*{Role}
You are an expert Python Developer and MCP (Model Context Protocol) Implementation Generator.
Your task is to produce a SINGLE, COMPLETE, EXECUTABLE Python file that implements class-based MCP tools using \texttt{mcp.server.fastmcp} with scenario-based state management and Pydantic schema validation.

\subsection*{CRITICAL OUTPUT RULES}
\begin{enumerate}[leftmargin=*]
    \item Output ONLY the final Python code wrapped in \texttt{<tool\_code>} tags.
    \item NO explanations, NO markdown formatting outside the tags.
    \item The code must be production-ready, strictly following the 4-Section structure.
\end{enumerate}

\subsection*{Implementation Architecture}

\subsubsection*{1. File Structure (Mandatory)}
\begin{itemize}[leftmargin=*]
    \item \textbf{Section 1: Schema}: Pydantic models (Entity models + 1 Scenario model). \texttt{Scenario\_Schema} defines the internal state structure of the Class.
    \item \textbf{Section 2: Class}: Main logic class.
    \item \textbf{Section 3: MCP Tools}: FastMCP registration + Wrappers.
    \item \textbf{Section 4: Entry Point}: \texttt{mcp.run()}.
\end{itemize}

\subsubsection*{2. Core Requirements}

\paragraph*{2.1. Pydantic Models}
\begin{itemize}[leftmargin=*]
    \item Use Pydantic v2 API throughout. Do NOT use deprecated v1 patterns, such as \texttt{.dict()}. Use \texttt{model\_dump()} instead.
    \item Define all data structures using Pydantic \texttt{BaseModel} classes.
    \item Import: \texttt{from pydantic import BaseModel, Field} and \texttt{from typing import Dict, List, Optional, Union, Any}.
    \item Each model must inherit from \texttt{BaseModel}, use \texttt{Field()} with descriptions/defaults, include type hints, and have docstrings.
    \item Entity model naming rule: Entity model class names and field names MUST NOT start with underscore. For example, use \texttt{Item} not \texttt{\_Item} for class name, and use \texttt{id} not \texttt{\_id} for field name.
    \item Simplify Nested Structures: For fields in the Scenario model that store complex nested dictionaries or variable schemas, such as configuration maps, weather patterns, or lookup tables, \textbf{use \texttt{Dict[str, Any]} or \texttt{dict}}. Do NOT use strict complex recursive types, such as \texttt{Dict[str, Dict[str, Union[str, float]]]}, to ensure robustness during scenario loading.
    \item Create individual entity models and one main scenario model defining the complete scenario structure.
    \item External data tables, also known as reference lookup data, should be defined directly as fields in the Scenario model using \texttt{Field(default=\{\{...\}\})} or \texttt{Field(default\_factory=lambda: \{\{...\}\})} with 10--20 entries.
    \item Current Time Management: All references to ``current time'', ``now'', or ``current date/time'' MUST be stored as string fields in the Scenario model. For example:
\end{itemize}

\begin{verbatim}
current_time: str = Field(
    ...,
    pattern=r"^\d{4}-\d{2}-\d{2}T\d{2}:\d{2}:\d{2}$",
    description="Current timestamp in ISO 8601 format"
)
\end{verbatim}

\begin{itemize}[leftmargin=*]
    \item NEVER use \texttt{datetime.now()} or any real-time functions. Access current time through scenario state variables, such as \texttt{self.current\_time}.
    \item End with \texttt{Scenario\_Schema = [Model1, Model2, ScenarioModel]} listing all Pydantic classes.
    \item \texttt{Scenario\_Schema} represents the internal state structure of the Class instance.
\end{itemize}

\paragraph*{Pydantic Model Validation}
You MUST use Pydantic \texttt{Field} constraints to enforce data validation at the model level. This is the PRIMARY validation mechanism.

\begin{itemize}[leftmargin=*]
    \item \textbf{Pattern Validation}: Use \texttt{Field(..., pattern=r"pattern")} for string fields with format requirements, such as codes, IDs, times, and dates. \textbf{IMPORTANT: Always use raw strings (\texttt{r""}) for regex patterns.}
    \begin{itemize}[leftmargin=*]
        \item Time fields: \texttt{pattern=r"\^{}([0-1]?[0-9]|2[0-3]):[0-5][0-9]\$"} for HH:MM format.
        \item ISO 8601: \texttt{pattern=r"\^{}\textbackslash\textbackslash d\{4\}-\textbackslash\textbackslash d\{2\}-\textbackslash\textbackslash d\{2\}T\textbackslash\textbackslash d\{2\}:\textbackslash\textbackslash d\{2\}:\textbackslash\textbackslash d\{2\}\$"} for timestamps, and \texttt{pattern=r"\^{}\textbackslash\textbackslash d\{4\}-\textbackslash\textbackslash d\{2\}-\textbackslash\textbackslash d\{2\}\$"} for dates.
    \end{itemize}
    \item \textbf{Range Constraints}: Use \texttt{Field(..., ge=0)} for non-negative numeric fields, \texttt{Field(..., ge=0, le=100)} for ranges, and \texttt{Field(..., gt=0)} for strictly positive values.
    \item \textbf{Validation Philosophy}: Prefer \texttt{Field} constraints for ALL format/range validation. Use \texttt{@field\_validator} or \texttt{@model\_validator} only when \texttt{Field} constraints are insufficient, such as cross-field validation. Pydantic v2 automatically validates when data is loaded via \texttt{load\_scenario}.
\end{itemize}

\paragraph*{2.2. Implementation Pattern}
\begin{itemize}[leftmargin=*]
    \item Create a Python class containing all MCP tools as public methods.
    \item Private methods, starting with \texttt{\_}, are helpers and not registered as MCP tools.
    \item After the class, create FastMCP instance and class instance.
    \item Register each public method as MCP tool using \texttt{@mcp.tool()} decorator with wrapper functions.
    \item Method signatures must exactly match \texttt{input\_schema} properties with correct types and names.
    \item Return values must precisely match \texttt{output\_schema} structure.
    \item \textbf{MCP tool wrapper function return type annotations should use simple types}, such as \texttt{-> str} or \texttt{-> dict}, NOT specific Pydantic model types. For example, avoid \texttt{-> Tweet}; use \texttt{-> dict} instead.
\end{itemize}

\paragraph*{2.3. Required Methods}

\subparagraph*{2.3.1. \texttt{\_\_init\_\_}}
\begin{itemize}[leftmargin=*]
    \item Initialize all state variables as class attributes with type hints.
    \item Do not set default values. They come from scenario loading.
\end{itemize}

\subparagraph*{2.3.2. \texttt{load\_scenario} (REQUIRED)}
\begin{itemize}[leftmargin=*]
    \item Signature: \texttt{def load\_scenario(self, scenario: dict) -> None:}
    \item Instantiate main scenario Pydantic model. For example, \texttt{scenario\_model = ScenarioModel(**scenario)}.
    \item Assign validated fields to class attributes. For example, \texttt{self.field = scenario\_model.field}.
    \item Pydantic handles type conversion automatically.
    \item Return \texttt{None} on success, with empty \texttt{output\_schema}.
\end{itemize}

\subparagraph*{2.3.3. \texttt{save\_scenario} (REQUIRED)}
\begin{itemize}[leftmargin=*]
    \item Signature: \texttt{def save\_scenario(self) -> dict:}
    \item Return dictionary containing all current state variables.
    \item Serialize Pydantic model instances using \texttt{model\_dump()} in Pydantic v2. Do NOT use \texttt{.dict()}, which is deprecated.
    \item The returned dictionary structure must exactly match the structure expected by \texttt{load\_scenario}, with the same field names and types.
\end{itemize}

\paragraph*{2.4. State Management}
\begin{itemize}[leftmargin=*]
    \item \textbf{State as Truth}: Class instance holds all state. State variables, such as \texttt{self.xxx}, serve as internal database tables/collections. All tools must operate directly on state variables. NEVER simulate external APIs, network requests, or create fake/mock data.
    \item \textbf{Scenario Loading}: Convert \texttt{dict -> Pydantic Model -> self.variables}. Pydantic models define structure; state variables store actual data as dicts/lists.
    \item \textbf{Anti-Lazy Logic}: Lookup tools MUST query the reference data from Schema fields, such as \texttt{self.xxxMap}. NEVER return hardcoded values like \texttt{return 100}. Access reference data through class attributes that correspond to Schema fields, such as \texttt{self.taxRatesMap}, not hardcoded values.
    \item \textbf{Time Management}: When tools need to access ``current time'' or ``now'', they MUST read from scenario state variables, such as \texttt{self.current\_time} or \texttt{self.current\_date}. NEVER use \texttt{datetime.now()}, \texttt{time.time()}, or any real-time functions. All time values are provided through the scenario and stored as strings.
    \item Perform CRUD operations on state variables: READ, WRITE, CREATE, UPDATE, and DELETE. Handle missing data: return empty results for reads and error dicts for operations requiring existing data. NEVER create fake data to fill responses.
\end{itemize}

\paragraph*{2.5. Reference Data Management}

\subparagraph*{Static Reference Data Pattern (for lookup tools)}
\begin{enumerate}[leftmargin=*]
    \item \textbf{Scenario Model}: Add reference data fields directly to the Scenario model as ordinary Pydantic fields. For example, \texttt{taxRatesMap: Dict[str, float] = Field(default=\{\{...\}\}, description="...")}. Use \texttt{Field(default=\{\{...\}\})} or \texttt{Field(default\_factory=lambda: \{\{...\}\})} to set default values containing 10--20 entries. These fields are just like other Scenario fields, with no special handling needed.
    \item \textbf{Class Init}: Initialize corresponding class attributes. For example, \texttt{self.taxRatesMap: Dict[str, float] = \{\{\}\}}.
    \item \textbf{Load Scenario}: Pydantic automatically handles default values. If scenario provides the field, use the provided value; otherwise, use the default value from \texttt{Field(default=...)}.
    \item \textbf{Tool Methods}: Access reference data directly through class attributes, such as \texttt{self.taxRatesMap}. NEVER return hardcoded values.
    \item \textbf{Save Scenario}: Return all fields in the dictionary, including reference data fields.
\end{enumerate}

\paragraph*{2.6. Random Number Generation (Reproducibility)}
\begin{itemize}[leftmargin=*]
    \item \textbf{Avoid random when possible}: Prefer deterministic logic based on state variables or input parameters. If random is necessary, add \texttt{random\_seed: Optional[int] = Field(default=None, description="Random seed for reproducible results")} to the Scenario model, initialize \texttt{self.random\_seed} in \texttt{\_\_init\_\_}, and use \texttt{random.seed(self.random\_seed)} in methods that need randomness.
\end{itemize}

\subsubsection*{3. MCP Tools}

\paragraph*{3.1. Error Handling \& Empty Output}
\begin{itemize}[leftmargin=*]
    \item \textbf{Class methods}: MUST NOT contain try-except blocks or error detection logic. Directly perform operations. Return normal results or \texttt{None}, for empty output. Let exceptions propagate naturally.
    \item \textbf{MCP wrapper functions}: MUST use try-except blocks for all error detection and handling.
    \begin{itemize}[leftmargin=*]
        \item \textbf{Simplified validation}: MCP wrapper functions should perform ONLY basic parameter existence checks, such as non-empty and non-None, and type checks using \texttt{isinstance}.
        \item \textbf{DO NOT duplicate format/range validation}: Rely on Pydantic model validation when data passes through \texttt{load\_scenario} or when creating model instances.
        \item \textbf{Business logic checks}: Perform ID existence checks, state-dependent validations, such as ``item not found'' or ``insufficient balance'', and other business logic validations.
        \item If class method returns \texttt{None}, meaning empty output, return success message string with \texttt{-> str}. Otherwise return result directly with \texttt{-> dict}.
    \end{itemize}
\end{itemize}

\textbf{Validation Responsibility Division:}
\begin{itemize}[leftmargin=*]
    \item \textbf{Pydantic Models (Primary)}: Handle ALL format validation, including pattern validation, range validation using \texttt{ge}, \texttt{gt}, \texttt{le}, and \texttt{lt}, and type conversion.
    \item \textbf{MCP Tool Wrappers (Secondary)}: Handle parameter existence, such as non-empty values, basic type checks using \texttt{isinstance}, and business logic validations, such as ID existence and state checks.
    \item When data is loaded via \texttt{load\_scenario(scenario: dict)}, Pydantic automatically validates all fields. There is no need to re-validate format/ranges in tool wrappers.
\end{itemize}

\paragraph*{3.2. Docstring Requirements (CRITICAL)}
\begin{itemize}[leftmargin=*]
    \item \textbf{Class methods (Section 2)}: Only require a single-line docstring describing what the method does.
    \item \textbf{MCP tool wrapper functions (Section 3)}: MUST have complete Google-style docstring with three sections:
    \begin{itemize}[leftmargin=*]
        \item \textbf{Description}: What the method does.
        \item \textbf{Args}: All parameters with types and descriptions. Mark optional with \texttt{[Optional]}.
        \item \textbf{Returns}: All return fields with types and descriptions, matching \texttt{output\_schema}.
    \end{itemize}
\end{itemize}

\subsection*{Reference Implementation}
\begin{verbatim}
{MCPToolGenerator_Example}
\end{verbatim}

\subsection*{Task}
Generate the MCP tool code based on the user's specific requirement.
Output ONLY the code inside \texttt{<tool\_code>...</tool\_code>}.

\end{tcolorbox}
\begin{tcolorbox}[
    breakable,
    colback=gray!20,
    colframe=gray!50,
    title=Test Cases Generation Prompt,
    fontupper=\small
]

\subsection*{Role}
You are an expert test scenario generator for MCP tool implementations. Your goal is to create diverse, comprehensive test scenarios that thoroughly validate tool functionality.

\subsection*{Responsibilities}

\subsubsection*{1. Analyze Tool Code Structure}
\begin{itemize}[leftmargin=*]
    \item Examine the provided \texttt{tool\_code} to identify the main Pydantic scenario model, such as \texttt{GoogleCalendarScenario}, \texttt{TwitterScenario}, or \texttt{InventoryScenario}.
    \item Understand all fields, their types, default values, and relationships.
    \item Identify reference data fields, such as lookup tables like \texttt{taxRatesMap} and \texttt{shippingZonesMap}.
    \item Understand the tool methods and their expected behaviors.
\end{itemize}

\subsubsection*{1.5 Pydantic Model Type Matching (CRITICAL)}
Before generating scenario data, you MUST carefully analyze the Pydantic model field types to ensure exact type matching.

\paragraph*{Complex Type Patterns}
\begin{itemize}[leftmargin=*]
    \item \textbf{\texttt{Dict[str, BaseModel]}}: Generate:
\end{itemize}

\begin{verbatim}
{"key": {"field1": value1, "field2": value2, ...}}
\end{verbatim}

Example:

\begin{verbatim}
tickets: Dict[str, TicketInfo]
\end{verbatim}

Should become:

\begin{verbatim}
{"T001": {"price": 100.0, "seats": 50}}
\end{verbatim}

WRONG:

\begin{verbatim}
{"T001": "ticket_info"}
\end{verbatim}

or:

\begin{verbatim}
{"T001": "some_string"}
\end{verbatim}

\begin{itemize}[leftmargin=*]
    \item \textbf{\texttt{List[BaseModel]}}: Generate:
\end{itemize}

\begin{verbatim}
[
  {"field1": value1, ...},
  {"field1": value2, ...}
]
\end{verbatim}

Example:

\begin{verbatim}
routes: List[RouteInfo]
\end{verbatim}

Should become:

\begin{verbatim}
[
  {"from": "BJ", "to": "SH"},
  {"from": "SH", "to": "GZ"}
]
\end{verbatim}

WRONG:

\begin{verbatim}
["route1", "route2"]
\end{verbatim}

or:

\begin{verbatim}
[{"name": "route1"}]
\end{verbatim}

\begin{itemize}[leftmargin=*]
    \item \textbf{Nested BaseModel classes}: Look for class definitions in the code.
\end{itemize}

For example, if you see:

\begin{verbatim}
class TicketInfo(BaseModel):
    price: float
    availability: int
\end{verbatim}

Then:

\begin{verbatim}
Dict[str, TicketInfo]
\end{verbatim}

expects:

\begin{verbatim}
{"key": {"price": 100.0, "availability": 50}}
\end{verbatim}

NOT:

\begin{verbatim}
{"key": {"ticket_id": "T001"}}
\end{verbatim}

because that uses wrong fields.

\begin{itemize}[leftmargin=*]
    \item \textbf{Type consistency}: Ensure value types match exactly.
    \begin{itemize}[leftmargin=*]
        \item \texttt{price: float} $\rightarrow$ Use \texttt{100.0} as a float, NOT \texttt{"100"} as a string. \texttt{100} as an integer is acceptable, but float is better.
        \item \texttt{count: int} $\rightarrow$ Use \texttt{50} as an integer, NOT \texttt{"50"} as a string.
        \item \texttt{available: bool} $\rightarrow$ Use \texttt{true/false} as boolean values, NOT \texttt{"true"} as a string.
    \end{itemize}
\end{itemize}

\paragraph*{Validation Rules}
\begin{enumerate}[leftmargin=*]
    \item Read ALL Pydantic class definitions in Section 1 (Schema) of the \texttt{tool\_code}.
    \item Map each field in the main Scenario model to its actual type.
    \item For complex types, such as \texttt{Dict} or \texttt{List} with \texttt{BaseModel}, identify the nested structure.
    \item Generate data that exactly matches the nested structure.
    \item DO NOT guess or simplify complex types. Match them precisely.
\end{enumerate}

\subsubsection*{2. Generate Diverse Test Scenarios}
You must generate \textbf{\texttt{\{n\_scenarios\}}} test scenarios with varying complexity levels.

\paragraph*{Complexity Levels}
\begin{enumerate}[leftmargin=*]
    \item \textbf{Simple (1--2 scenarios)}: Minimal data.
    \begin{itemize}[leftmargin=*]
        \item 1--2 main entities, such as 1 calendar with 1 event, or 2 items in inventory.
        \item Basic fields populated.
        \item Use default reference data if applicable.
        \item Purpose: Test basic tool functionality.
    \end{itemize}

    \item \textbf{Medium (2--3 scenarios)}: Moderate data.
    \begin{itemize}[leftmargin=*]
        \item 3--5 main entities with varied attributes.
        \item Mix of populated and empty optional fields.
        \item Some edge cases, such as events at midnight or items with zero price.
        \item Purpose: Test typical use cases.
    \end{itemize}

    \item \textbf{Complex (1--2 scenarios)}: Rich data.
    \begin{itemize}[leftmargin=*]
        \item 5--10 main entities with diverse relationships.
        \item All fields populated with realistic values.
        \item Nested structures fully used.
        \item Purpose: Test scalability and complex interactions.
        \item \textbf{IMPORTANT}: Focus on functional coverage, not data volume. Use representative data samples rather than exhaustive datasets to avoid JSON serialization/deserialization issues.
    \end{itemize}

    \item \textbf{Boundary (as needed)}: Edge cases.
    \begin{itemize}[leftmargin=*]
        \item Empty collections, such as no calendars or no items. These should pass.
        \item Extreme values, such as very long strings or max integers. These should pass or fail appropriately.
        \item \textbf{Invalid inputs}, such as special characters violating validation rules. These should be rejected.
        \item Purpose: Test error handling and edge cases.
    \end{itemize}
\end{enumerate}

\textbf{IMPORTANT}: For boundary scenarios that test invalid inputs, you MUST specify \texttt{expected\_behavior}:
\begin{itemize}[leftmargin=*]
    \item \texttt{"pass"}: Normal success expected, such as empty collections or extreme but valid values.
    \item \texttt{"validation\_error"}: Tool should reject input with validation error, such as special characters violating a pattern.
    \item If not specified, defaults to \texttt{"pass"}.
\end{itemize}

\subsubsection*{3. Ensure Scenario Quality}
Each scenario must:
\begin{itemize}[leftmargin=*]
    \item Be a complete, valid dictionary matching the scenario model structure.
    \item Include ALL required fields from the Pydantic model.
    \item Use realistic, coherent data, such as consistent date ranges and related IDs.
    \item Have unique identifiers, such as different event IDs or item IDs.
    \item Include reference data fields with their default values, or variations if testing lookup functionality.
    \item For boundary scenarios with invalid data, include \texttt{expected\_behavior: "validation\_error"} to indicate expected rejection.
    \item If the scenario model includes \texttt{random\_seed}, provide a fixed integer value, such as \texttt{42}, to ensure reproducible results.
\end{itemize}

\textbf{Data Volume Guidelines}:
\begin{itemize}[leftmargin=*]
    \item Keep scenario data concise and manageable to avoid JSON serialization/deserialization errors.
    \item For tools with large datasets, such as train schedules or route maps, use small but representative samples, usually 2--5 entries, rather than exhaustive data.
    \item Complex nested structures should be simplified. Test functionality, not data volume.
    \item If a tool involves lookup tables or reference data, include only essential entries needed for testing, typically 3--10 entries.
\end{itemize}

\subsubsection*{4. Output Format}
Your response must strictly follow this structure:

\begin{verbatim}
<scenarios>
[
  {
    "scenario_id": "scenario_001",
    "complexity_level": "simple",
    "description": "Brief description of what this scenario tests",
    "expected_behavior": "pass",
    "scenario_data": {
      // Complete scenario dictionary matching the Pydantic model
    }
  },
  {
    "scenario_id": "scenario_002",
    "complexity_level": "medium",
    "description": "Brief description of what this scenario tests",
    "expected_behavior": "pass",
    "scenario_data": {
      // Complete scenario dictionary matching the Pydantic model
    }
  },
  {
    "scenario_id": "scenario_005",
    "complexity_level": "boundary",
    "description": "Test invalid data with special characters (should be rejected)",
    "expected_behavior": "validation_error",
    "scenario_data": {
      // Scenario with intentionally invalid data
    }
  }
  // ... more scenarios up to {n_scenarios}
]
</scenarios>
\end{verbatim}

\subsection*{Important Notes}
\begin{itemize}[leftmargin=*]
    \item Each \texttt{scenario\_id} must be unique, such as \texttt{"scenario\_001"} or \texttt{"scenario\_002"}.
    \item \texttt{complexity\_level} must be one of: \texttt{"simple"}, \texttt{"medium"}, \texttt{"complex"}, or \texttt{"boundary"}.
    \item \texttt{expected\_behavior} must be \texttt{"pass"} by default or \texttt{"validation\_error"} for scenarios testing invalid input rejection.
    \item \texttt{scenario\_data} must be a complete, valid scenario dictionary.
    \item Include variety in your test data to maximize test coverage.
\end{itemize}

\end{tcolorbox}
\begin{tcolorbox}[
    breakable,
    colback=gray!20,
    colframe=gray!50,
    title=Validation Prompt,
    fontupper=\small
]

\subsection*{Role}
You are a comprehensive MCP tool validator. Your task is to validate a single test scenario by executing all available tools and diagnosing any issues.

\subsection*{Responsibilities}

\subsubsection*{1. Scenario Preparation}
You will receive:
\begin{itemize}[leftmargin=*]
    \item \texttt{mcp\_server\_name}: Name of the MCP server
    \item \texttt{tool\_code}: MCP Tools section (Section 3) containing FastMCP registration and tool wrapper functions
    \item \texttt{tools\_metadata}: List of all available tools with their schemas
    \item \texttt{scenario\_id}: Unique identifier for this scenario
    \item \texttt{scenario\_data}: The test scenario data
    \item \texttt{request\_id}: For constructing \texttt{client\_id}
\end{itemize}

\subsubsection*{2. Client ID Construction}
You must use this exact pattern:
\begin{itemize}[leftmargin=*]
    \item \texttt{client\_id = "\{mcp\_server\_name\}-\{request\_id\}\_\{scenario\_id\}"}
    \item Example: \texttt{"GoogleMaps-abc123\_scenario\_001"}
    \item Use the SAME \texttt{client\_id} for all operations in this scenario
\end{itemize}

\subsubsection*{3. Understanding Expected Behavior}
The scenario may include an \texttt{expected\_behavior} field:
\begin{itemize}[leftmargin=*]
    \item \textbf{\texttt{"pass"}} (default): Normal execution, tools should succeed
    \item \textbf{\texttt{"validation\_error"}}: Scenario contains invalid data, tools should reject it with validation error
\end{itemize}

When evaluating results:
\begin{enumerate}[leftmargin=*]
    \item \textbf{Pass}: Tool executed successfully with expected output when \texttt{expected\_behavior="pass"}.
    \item \textbf{Expected Failure}: Tool correctly rejected invalid input with validation error when \texttt{expected\_behavior="validation\_error"}.
    \begin{itemize}[leftmargin=*]
        \item THIS COUNTS AS PASSED. The tool is working correctly by rejecting bad data.
    \end{itemize}
    \item \textbf{Unexpected Failure}:
    \begin{itemize}[leftmargin=*]
        \item Tool raised error when success was expected, meaning \texttt{expected\_behavior="pass"} but got error.
        \item OR: Tool succeeded when validation error was expected, meaning \texttt{expected\_behavior="validation\_error"} but no error.
    \end{itemize}
\end{enumerate}

\subsubsection*{4. Layered Validation Procedure}

\paragraph*{Layer 1: Scenario Loading (Critical and Blocking)}
Call \texttt{execute\_mcp\_tool} with:
\begin{itemize}[leftmargin=*]
    \item \texttt{tool\_name}: \texttt{"\{mcp\_server\_name\}-load\_scenario"}
    \item \texttt{tool\_args}: \texttt{\{"scenario": scenario\_data\}}
    \item \texttt{client\_id}: as constructed above
\end{itemize}

Record the result.

Evaluate based on \texttt{expected\_behavior}:
\begin{itemize}[leftmargin=*]
    \item If \texttt{expected\_behavior="validation\_error"} and \texttt{load\_scenario} fails with validation error: PASS, as an expected failure.
    \item If \texttt{expected\_behavior="pass"} and \texttt{load\_scenario} succeeds: PASS.
    \item Otherwise: FAIL, as unexpected behavior.
\end{itemize}

\textbf{IMPORTANT}: If \texttt{load\_scenario} FAILS unexpectedly:
\begin{itemize}[leftmargin=*]
    \item Mark it as CRITICAL error in the errors list.
    \item \textbf{STOP validation immediately and return}. Do not proceed to test other tools.
    \item This is a blocking failure that prevents further validation.
\end{itemize}

\paragraph*{Layer 2: Tool Execution (Conditional)}
\textbf{Only execute if \texttt{load\_scenario} succeeded}:

For each tool in \texttt{tools\_metadata}, excluding \texttt{load\_scenario} and \texttt{save\_scenario}:
\begin{itemize}[leftmargin=*]
    \item Use the loaded scenario state.
    \item Design 2--3 test cases with different inputs:
    \begin{itemize}[leftmargin=*]
        \item \textbf{Valid case}: Normal, expected inputs
        \item \textbf{Boundary case}: Edge values, if applicable
        \item \textbf{Error case}: Invalid inputs, if error handling should be tested
    \end{itemize}
    \item For each test case:
    \begin{itemize}[leftmargin=*]
        \item Call \texttt{execute\_mcp\_tool} with the tool and test inputs
        \item Record: input, expected behavior, actual output, and any errors
        \item Evaluate: Does output match expected? Are there any unexpected errors?
    \end{itemize}
    \item This layer helps identify tool logic errors independent of scenario loading.
\end{itemize}

\paragraph*{Layer 3: State Consistency (Conditional)}
\begin{itemize}[leftmargin=*]
    \item Only run this if \texttt{load\_scenario} succeeded.
    \item After executing all tools, call:
    \begin{itemize}[leftmargin=*]
        \item \texttt{tool\_name}: \texttt{"\{mcp\_server\_name\}-save\_scenario"}
        \item \texttt{tool\_args}: \texttt{\{\{\}\}}
        \item \texttt{client\_id}: same as before
    \end{itemize}
    \item Record the saved scenario.
    \item Compare with the original scenario + expected modifications.
    \item If \texttt{load\_scenario} failed, skip this step with note \texttt{"Skipped due to load\_scenario failure"}.
\end{itemize}

\subsubsection*{4. Error Diagnosis}
For any failures, provide:
\begin{itemize}[leftmargin=*]
    \item \textbf{Error type}: For example, \texttt{"Tool execution error"}, \texttt{"State inconsistency"}, or \texttt{"Schema mismatch"}
    \item \textbf{Error location}: Which tool/method failed
    \item \textbf{Error details}: Actual error message, stack trace if available
    \item \textbf{Expected vs Actual}: What was expected vs what happened
    \item \textbf{Root cause analysis}: Why did this fail? For example, \texttt{"load\_scenario does not handle empty lists"} or \texttt{"tool returns wrong field name"}
\end{itemize}

\subsubsection*{6. Output Format}
Your response must strictly follow:

\begin{verbatim}
<validation_result>
{
  "scenario_id": "...",
  "passed": true/false,
  "load_scenario_result": {
    "success": true/false,
    "error": "..."  // provide the error message if failed
  },
  "tool_execution_results": [
    {
      "tool_name": "...",
      "passed": true/false,
      "error": "..."  // provide the error message if failed
    }
  ],
  "save_scenario_result": {
    "success": true/false,
    "consistency_check": true/false,
    "error": "..."  // provide the error message if failed
  },
  "errors": [
    {
      "error_type": "...",
      "error_location": "...",
      "error_details": "...",
      "expected_vs_actual": "...",
      "root_cause": "...",
      "expected_error": true/false
    }
  ]
}
</validation_result>
\end{verbatim}

\subsection*{Important Notes}
\begin{itemize}[leftmargin=*]
    \item Test ALL tools, except \texttt{load\_scenario} and \texttt{save\_scenario}.
    \item Use the SAME \texttt{client\_id} throughout.
    \item Properly classify \texttt{result\_type} based on \texttt{expected\_behavior} from the scenario.
    \item For \texttt{expected\_behavior="validation\_error"}, set \texttt{expected\_error=true} when validation error occurs.
    \item Provide detailed error diagnosis for UNEXPECTED failures only.
    \item Even if one tool fails, continue testing other tools.
\end{itemize}

\end{tcolorbox}
\begin{tcolorbox}[
    breakable,
    colback=gray!20,
    colframe=gray!50,
    title=Tool Revise Prompt,
    fontupper=\small
]

\subsection*{Role}
You are an expert MCP tool code reviser. Your task is to analyze validation failures from multiple test scenarios and fix all issues systematically.

\subsection*{Core Responsibilities}

\subsubsection*{1. Error Categorization and Scenario Problem Detection}
Categorize errors into:
\begin{itemize}[leftmargin=*]
    \item \textbf{Pydantic Model Issues}: Schema definition problems, field type mismatches
    \item \textbf{Load/Save Scenario Issues}: State management problems, missing fields in save
    \item \textbf{Tool Logic Errors}: Incorrect implementation, wrong return values, missing error handling
    \item \textbf{State Management Issues}: Tools not reading/writing state correctly, state inconsistencies
    \item \textbf{Schema Mismatches}: Input/output does not match declared schemas
\end{itemize}

\textbf{IMPORTANT: Distinguish between Code Problems and Scenario Problems}

Before fixing code, you must determine if the failures are due to:
\begin{itemize}[leftmargin=*]
    \item \textbf{Scenario Problems} (\texttt{is\_scenario\_problem=true}): Failures caused by invalid or poorly designed test scenarios
    \begin{itemize}[leftmargin=*]
        \item Scenario data does not match tool schema, such as missing fields, wrong types, or values out of range
        \item Scenario data has logical errors, such as references to non-existent IDs or inconsistent data
        \item Scenario's \texttt{expected\_behavior} is incorrectly set, such as should be \texttt{"pass"} but marked as \texttt{"validation\_error"}, or vice versa
        \item Scenario tests non-existent functionality or tools
        \item Scenario violates tool's documented constraints or requirements
    \end{itemize}

    \item \textbf{Code Problems} (\texttt{is\_scenario\_problem=false}): Failures caused by tool implementation issues
    \begin{itemize}[leftmargin=*]
        \item Tool logic errors in implementation
        \item Incomplete or incorrect schema definitions
        \item State management problems
        \item Missing error handling
        \item Tools not following MCP requirements
    \end{itemize}
\end{itemize}

\textbf{Judgment Principle}: If errors are due to scenario data not meeting tool requirements or poor scenario design, set \texttt{is\_scenario\_problem=true}. If errors are due to tool code implementation issues, set \texttt{is\_scenario\_problem=false}.

\subsubsection*{2. Prioritized Fix Strategy}
The errors are categorized by severity. Fix issues in this order:

\begin{enumerate}[leftmargin=*]
    \item \textbf{CRITICAL (Must Fix First)}:
    \begin{itemize}[leftmargin=*]
        \item \texttt{load\_scenario} failures, since these block all testing
        \item Pydantic model schema mismatches, including validation errors and type mismatches
        \item These affect all scenarios and must be fixed before anything else
    \end{itemize}

    \item \textbf{HIGH (Fix Next)}:
    \begin{itemize}[leftmargin=*]
        \item Tools that fail in multiple scenarios
        \item Tool logic errors affecting core functionality
        \item State management issues
    \end{itemize}
\end{enumerate}

\textbf{IMPORTANT}: If the error severity summary shows CRITICAL errors with count greater than 0, you MUST fix those first before addressing other issues. Do not make changes to working code until critical issues are resolved.

\subsubsection*{3. Fix Implementation Guidelines}
\begin{itemize}[leftmargin=*]
    \item Fix the root cause, not symptoms
    \item Ensure fixes do not break currently passing scenarios
    \item Maintain all original functionality and structure
    \item Follow all MCP tool generation requirements
    \item Test edge cases in your mental model before suggesting fixes
\end{itemize}

When fixing, verify:
\begin{itemize}[leftmargin=*]
    \item All Pydantic models are complete and correct
    \item \texttt{load\_scenario} properly validates and assigns all fields
    \item \texttt{save\_scenario} returns all current state fields
    \item Tool methods correctly access state via \texttt{self.xxx}
    \item Return values exactly match output schemas
    \item Error handling for missing data and invalid inputs
    \item Reference data fields are properly initialized and used
\end{itemize}
\end{tcolorbox}
\subsection{Prompts for ToolGraph}
\begin{tcolorbox}[
    breakable, 
    colback=gray!20, 
    colframe=gray!50, 
    title=Logical Refinement Prompt for ToolGraph,
    fontupper=\small
]
\subsection*{Role}
You are an expert tool relationship analyst specializing in dependency inference.
Your task is to \textbf{augment} the current tool dependency graph by adding \textit{only missing, justified directed edges}.

You are given:
\begin{enumerate}[leftmargin=*]
    \item \textbf{Tool Descriptions}: A list of tools, each with name, functional description, and parameters.
    \item \textbf{Current Adjacency Map}: A dict \texttt{tool\_name $\rightarrow$ [list of successor tool names]}, representing \textit{existing} dependencies (Tool A $\rightarrow$ Tool B means Tool B may depend on or follow Tool A).
\end{enumerate}

\subsection*{Guidelines}
For every \textit{candidate} ordered pair (Tool A $\rightarrow$ Tool B) \textbf{not already present}, assess:
\begin{itemize}[leftmargin=*]
    \item \textbf{Semantic Complementarity}: Do the tools solve parts of a shared task or pipeline? (e.g., preprocessing $\rightarrow$ analysis)
    \item \textbf{Data Flow Feasibility}: Can outputs (explicit, implicit, or inferred context) from Tool A reasonably inform or enable Tool B's execution?
    \item \textbf{Workflow Plausibility}: Would a rational user \textit{naturally} run Tool B after Tool A in a realistic scenario?
    \item \textbf{Parameter/Context Alignment}: Are parameters, domains, or expected inputs/outputs conceptually aligned---even if naming differs?
\end{itemize}

\subsection*{Constraints}
\begin{itemize}[leftmargin=*]
    \item No self-loops (Tool A $\rightarrow$ Tool A is forbidden).
    \item Only add edges where Tool A and Tool B are distinct and exist in the tool list.
\end{itemize}
\end{tcolorbox}
\subsection{Prompts for QueryGen}
\begin{tcolorbox}[
    breakable, 
    colback=gray!20, 
    colframe=gray!50, 
    title=Simulated User Prompt,
    fontupper=\small
]
\subsection*{Role}
You are a realistic human user interacting naturally with an assistant.

\subsection*{Here is what you know:}
\begin{itemize}[leftmargin=*]
    \item \textbf{Scenario and User Profile:}\begin{verbatim}{scenario}\end{verbatim}
    \item \textbf{User Intent:}\begin{verbatim}{user_intent}\end{verbatim}
    \item \textbf{Hidden MCP Servers:}\begin{verbatim}{mcp_server_config}\end{verbatim}
\end{itemize}

\subsection*{Here are the available tools you can use:}
You have access to the following tools within \texttt{<tools></tools>} XML tags:
\begin{verbatim}
<tools>
{mcp_server_tools}
</tools>
\end{verbatim}

\subsection*{Conversation}
\begin{verbatim}
{conversation}
\end{verbatim}

\subsection*{Guidelines}
Your response should follow these guidelines step-by-step:
\begin{enumerate}[leftmargin=*]
    \item \textbf{Natural Voice}
    \begin{itemize}
        \item Speak in first-person as a real person would speak. Respond conversationally and naturally.
    \end{itemize}

    \item \textbf{Task Scope}
    \begin{itemize}
        \item Focus only on the current turn. When the assistant asks if you have anything else, do not invent new requests.
    \end{itemize}

    \item \textbf{Knowledge Boundaries}
    \begin{itemize}
        \item Only share knowledge a real person would realistically recall, as indicated in your user knowledge.
    \end{itemize}

    \item \textbf{Don't Over-Help}
    \begin{itemize}
        \item If the information the assistant is asking for is already provided in previous conversation or should be discovered through tools or reasoning, do not directly provide. Instead, hint to the assistant how to get it.
        \item Don't describe what you'll do. Directly and concisely respond to the assistant's question.
    \end{itemize}

    \item \textbf{Tool-Use Discipline}
    \begin{itemize}
        \item When the assistant gives you a direct, actionable instruction, you may use tools:
        \begin{itemize}
            \item "Can you login to the website with your account?"
            \item "Try restarting the car engine."
            \item "Please turn on your mobile phone."
        \end{itemize}
        \item Only use tools when explicitly instructed and actionable. Never use tools for inquiry or general questions.
        \item You can only use tools from the "Available User Tools" section.
        \item After using tools, you need to respond to the assistant accordingly.
    \end{itemize}
\end{enumerate}

\subsection*{Response Format}
For each function call, return a JSON object with function name and arguments within \texttt{<tool\_call></tool\_call>} XML tags:
\begin{verbatim}
<tool_call>
{"name": <function-name>, "arguments": <args-json-object>}
</tool_call>
\end{verbatim}

Otherwise, respond to the assistant directly. Please check each guideline before responding. Please avoid calling tools and responding to the assistant in the same step.
\end{tcolorbox}
\begin{tcolorbox}[
    breakable, 
    colback=gray!20, 
    colframe=gray!50, 
    title=Assistant Prompt,
    fontupper=\small
]
\subsection*{Role}
You are a helpful assistant. Your goal is to fulfill the user's requests in an interactive environment.

At each step, you will receive either the user's request/reply or the tool call results.
\begin{itemize}[leftmargin=*]
    \item If you can proceed with the current information, select proper tools from the tool set and provide complete, valid parameters.
    \item If you lack essential information to complete the task or perform a tool call, and it cannot be obtained through the existing tool set, actively ask the user for specific details.
    \item Avoid calling tools while interacting with user in one step.
    \item When a task involves sensitive credentials or physical device actions (e.g., logging into an account or restarting a phone), provide explicit step-by-step instructions naming the specific tools and required parameters.
    \item You cannot execute user tools directly; instead, guide users on how to perform these actions themselves.
    \item When you believe the task is completed, provide a direct and concise response to the user's original request.
\end{itemize}

Here are the actions you may instruct the user to do:
\begin{verbatim}
{user_tools}
\end{verbatim}

\subsection*{Conversation}
\begin{verbatim}
{conversation}
\end{verbatim}

\subsection*{Tools}
You may call one or more functions to assist with the user query.

You are provided with function signatures within \texttt{<tools></tools>} XML tags:
\begin{verbatim}
<tools>
{tools}
</tools>
\end{verbatim}

For each function call, return a JSON object with function name and arguments within \texttt{<tool\_call></tool\_call>} XML tags:
\begin{verbatim}
<tool_call>
{"name": <function-name>, "arguments": <args-json-object>}
</tool_call>
\end{verbatim}
\end{tcolorbox}

\begin{tcolorbox}[
    breakable, 
    colback=gray!20, 
    colframe=gray!50, 
    title=Scenario Planner Prompt,
    fontupper=\small
]
\subsection*{Role}
You are a scenario planner specialized in creating high-quality initial contextual scenario for multi-turn conversation.
You will be given:
\begin{enumerate}[leftmargin=*]
    \item \textbf{Tool Call Trace:} A list of tools (from various MCP servers) that appeared in a conversation.
\end{enumerate}

\subsection*{Instruction}
\subsubsection*{1. Scenario Design}
Design a cohesive narrative that \textit{naturally motivates} the observed tool sequence. Your scenario must:
\begin{itemize}[leftmargin=*]
    \item Define a realistic user persona (name, age range, occupation, location, relevant traits)
    \item Establish a concrete situation with time/place/context that explains \textit{why} the user would perform these actions
    \item Flow logically from initial need $\rightarrow$ actions taken $\rightarrow$ implied next steps
    \item \textbf{Never mention tools, APIs, or technical mechanisms}---describe only human behaviors and motivations
    \item Be specific and grounded (avoid generic phrases like ``a user wanted information'')
    \item Reflect cultural and situational plausibility for the geographic/occupational context
\end{itemize}
\end{tcolorbox}
\begin{tcolorbox}[
    breakable, 
    colback=gray!20, 
    colframe=gray!50, 
    title=Query Generation Prompt,
    fontupper=\small
]
\subsection*{Role}
You are a simulated user. Your task is to generate the most plausible, natural user request that would directly and exclusively motivate the target tool call(s) in the current turn.

\subsection*{Guidelines}
\subsubsection*{Clarity \& Naturalness}
\begin{itemize}[leftmargin=*]
    \item Be conversational and realistic. Avoid robotic phrasing, checklists, overly technical jargon, or specific tool name and parameter.
    \item Speak strictly in the first-person perspective.
    \item Build logically on prior context using natural references (e.g., ``the hotel you found earlier'', ``since we're sticking to that budget'').
\end{itemize}

\subsubsection*{Background Analysis}
\begin{itemize}[leftmargin=*]
    \item Analyze what previous turns accomplished.
    \item Use the scenario and user profile to shape tone, preferences, and constraints (e.g., budget-conscious, eco-friendly).
\end{itemize}

\subsubsection*{Target Tool Analysis}
\begin{itemize}[leftmargin=*]
    \item Analyze the provided target tool calls to identify their underlying subgoals.
    \item Determine the logical relationship between these subgoals (sequential, parallel, or conditional).
    \item Weave them into a single, cohesive natural language query that naturally motivates all tool executions.
    \item Ensure smooth transitions and logical flow, concatenating independent subgoals where appropriate.
\end{itemize}

\subsubsection*{User Intent}
\begin{itemize}[leftmargin=*]
    \item Firstly briefly analyze what previous turns achieve, then explain the user intent for this turn, including the goals, constraints (e.g. tight budget), and the preferences (e.g. cheaper ticket).
\end{itemize}
\end{tcolorbox}

\end{document}